\documentclass{article}

\usepackage[english]{babel}

\usepackage[letterpaper,top=2cm,bottom=2cm,left=3cm,right=3cm,marginparwidth=1.75cm]{geometry}

\usepackage{natbib}
\usepackage{amsmath}
\usepackage{amssymb}
\usepackage{mathrsfs}
\usepackage{graphicx}
\usepackage{tabularx} 
\usepackage{subcaption}
\usepackage[colorlinks=true, allcolors=blue]{hyperref}

\title{Evaluating Contextual Bias in CNN Image Classification: Evidence from Agricultural Benchmark Datasets}
\author{Abhilekha Dalal, Michael Okonoda, Eder Martinez, Lior Shamir}

\begin{document}
\maketitle

\begin{abstract}
Convolutional neural networks (CNNs) are typically evaluated using held-out classification accuracy, an approach that presupposes predictions are based primarily on the intended object of interest rather than incidental surrounding context. We test this assumption in CNN-based agricultural image classification by comparing model performance on original images with performance on background-dominated patches extracted from the same images across eight publicly available agricultural benchmark datasets and four widely used CNN architectures. Background-dominated patches were classified above dataset-specific random chance for six of the eight datasets, and substantially above chance for four of them, indicating that contextual information contributes to model predictions for the majority of datasets evaluated. For these four datasets, we further evaluated whether this behavior reflected genuine class-discriminative information or was primarily attributable to class imbalance using macro-averaged precision, recall, and F1 together with class-balanced test subsets. The results show that contextual reliance does not admit a single explanation: class imbalance accounts for a substantial portion of the observed signal for some datasets and architectures, whereas above-chance contextual classification persists after balancing for others. Together with previous evidence from curated object recognition and cancer pathology imaging, these findings support the growing view that contextual bias is a recurring characteristic of CNN-based image classification rather than a phenomenon confined to a single application domain. More broadly, this work provides a systematic framework for quantifying contextual bias across heterogeneous image datasets by combining dataset-specific random-chance baselines, contextual bias categorization, macro-averaged evaluation, and class-balanced robustness analysis.

\end{abstract}


\section{Introduction}
\label{intro}
Convolutional neural networks (CNNs) have become the dominant approach for image classification and have achieved remarkable performance across a wide range of computer vision tasks. Their success has led to widespread adoption in domains ranging from image classification~\cite {ramprasath2018image}, object recognition~\citep{hary2022object}, drug design~\cite{segler2018generating}, medical image analysis~\cite{choi2019artificial} to precision agriculture~\citep{hasan2021survey,ramcharan2017deep}.
Despite these successes, CNNs remain inherently black-box models, making it difficult to determine what visual evidence they learn from training data and subsequently use during prediction~\citep{black_box2020,brown2020language,shahroudnejad2021survey}.

Since CNNs learn statistical patterns directly from training data, they may exploit contextual correlations introduced during data collection (e.g., consistent backgrounds, lighting conditions, or imaging artifacts) or other correlated visual cues that enable accurate predictions without relying on the target object itself. Consequently, model evaluation typically relies on empirical performance measures such as classification accuracy, confusion matrices, F1-score, etc. Although these metrics quantify predictive performance, they provide little insight into whether a model has learned meaningful visual characteristics intended for the task or has instead relied on contextual correlations present in the training data. As a result, high benchmark performance may overestimate a model's ability to generalize to new environments, and as these models are increasingly deployed in real-world applications, it is equally important to understand the visual evidence that CNNs rely on during prediction.

This observation has motivated a growing body of work investigating contextual bias in CNN-based image classification~\citep{shetty2019not,ribeiro2016should,zech2018variable}. Earlier studies reported dataset bias in object recognition benchmarks, face recognition datasets, and astronomical image analysis by showing that CNNs often exploit contextual information or dataset-specific characteristics during prediction~\citep{dhar2022systematic,shamir2008evaluation,beery2018recognition}. More recently, studies on synthetically generated image datasets and cancer pathology images demonstrated that CNNs can correctly classify images using background information at accuracies substantially above random chance~\citep{okonoda2026unmasking,martinez2026detection}. Although these studies examine visually and semantically unrelated domains, they consistently report the same underlying phenomenon: CNNs can exploit contextual cues that remain predictive of the class labels. Collectively, these findings suggest that contextual bias may reflect a broader characteristic of CNN-based image classification rather than a property of individual datasets or application domains.

Despite this growing body of evidence, existing studies remain fragmented. Most investigations focus on individual datasets within a single application domain, making it difficult to determine whether the observed behavior reflects properties of specific datasets or persists across more heterogeneous visual data~\citep{nene1996columbia,shamir2008evaluation,mohanty2016using,martinez2026detection}. In particular, few studies have examined contextual bias using collections of datasets that differ substantially in imaging conditions, acquisition protocols, numbers of classes, and semantic content while applying a consistent evaluation methodology. Such analyses are necessary to understand whether contextual bias is an isolated phenomenon or a recurring property of CNN-based image classification.

Agricultural image classification provides an ideal setting to investigate this question. In agriculture, CNNs support applications such as crop disease diagnosis, weed identification, plant species recognition, and automated crop monitoring, where accurate image classification assists precision agriculture and large-scale decision-making~\citep {ramcharan2017deep,ferentinos2018deep,hasan2021survey,rai2023applications,kwak2019impact}. Also, agricultural datasets encompass a wide range of crops, weeds, plant diseases, imaging environments, and acquisition protocols, ranging from controlled laboratory conditions to highly variable field environments. These differences introduce substantial variability in background appearance, illumination, surrounding vegetation, soil characteristics, and other contextual information that may become correlated with class labels. Previous agricultural studies have reported isolated observations indicating that CNNs may rely on crop identity, healthy plant regions, or background characteristics instead of disease symptoms or crop-specific traits~\citep{mohanty2016using,atabay2017,ferentinos2018deep,toda2019convolutional,lee2020new}. However, a systematic evaluation across diverse agricultural datasets has not yet been conducted, leaving unanswered whether contextual bias consistently emerges across heterogeneous agricultural image classification benchmarks.

To address these questions, we compare CNN performance on original agricultural images with performance on background-dominated image patches that contain little or no visual information about the primary object of interest. If CNNs primarily learn object-specific visual representations, classification from these background regions should not substantially exceed dataset-specific random chance accuracy. Conversely, consistently above-chance performance indicates that contextual information contributes to the learned representations and influences model predictions. This study is guided by the following research questions:

\textbf{RQ1.} \textit{Can CNN architectures trained for agricultural image classification correctly classify images from background-dominated regions at accuracies exceeding dataset-specific random chance?}\\

\textbf{RQ2.} \textit{How does contextual reliance vary across agricultural datasets and CNN architectures, and does the observed behavior persist after accounting for class imbalance through macro-averaged and class-balanced evaluation?}\\

This study evaluates eight widely used agricultural benchmark datasets together with four commonly used CNN architectures using a unified evaluation methodology. Because these datasets differ considerably in the number of classes, imaging conditions, and acquisition protocols, we adopt a comparative evaluation framework based on dataset-specific random chance accuracy, patch classification performance, and robustness analyses using macro-averaged and balanced evaluation metrics. Our analysis demonstrates that several CNNs achieve substantially above-chance classification using background-dominated image regions and, in some cases, approach the performance obtained from the original images. When considered alongside previous findings in object recognition, cancer pathology, and generated image datasets, our results provide the strongest evidence to date that contextual bias is a recurring characteristic of CNN-based image classification rather than an isolated property of a particular application domain. These findings highlight the need to complement conventional benchmark evaluation with empirical assessments of contextual bias in order to better understand the reliability and generalizability of CNN-based image classification systems.

\section{Related Work}
\label{literature}


The recognition that convolutional neural networks (CNNs) can exploit information beyond the intended object of interest emerged gradually through evidence accumulated across diverse computer vision domains. The concept of dataset bias was formalized by ~\citep{torralba2011unbiased}, who demonstrated that images could be classified according to the benchmark dataset from which they originated, revealing that datasets possess distinctive acquisition signatures unrelated to object identity. Subsequent studies showed that this behavior extends beyond dataset-specific acquisition characteristics. ~\citep{geirhos2018imagenet} formalized this pattern as shortcut learning - models exploiting statistical regularities that are predictive on the training distribution but not aligned with the intended task, demonstrating that ImageNet-trained CNNs preferentially relied on local texture rather than global object shape
while ~\citep{beery2018recognition} reported that wildlife classifiers generalized poorly to unseen camera locations because background and location characteristics had become predictive of species identity. Similar observations were reported in face recognition~\citep{banerjee2023analyzing} and medical imaging~\citep{zech2018variable}, where CNNs exploited surrounding context or acquisition-specific artifacts rather than relying exclusively on the intended visual features. Collectively, these studies demonstrated that CNNs frequently exploit contextual and dataset-specific cues beyond the primary object of interest. The evidence supporting this conclusion, however, was consistently indirect - context ablation, saliency visualization, among other indirect diagnostic techniques, motivating the need for methodologies capable of directly quantifying contextual reliance. 

A more direct approach for quantifying contextual reliance is to evaluate CNNs on image regions from which the primary object of interest has been removed. ~\citep{shamir2008evaluation} classified face-dataset images using only small background regions excluding the face, hair, and clothing, and found several benchmarks remained classifiable well above random chance. Similar observations were later reported for fluorescence microscopy images~\citep{shamir2011assessing}, where seemingly non-informative image regions revealed systematic dataset biases affecting automated image analysis. ~\citep{model2015comparison} extended this methodology to object recognition benchmarks by extracting visually uninformative image patches and showed that all evaluated datasets permitted above-chance classification despite the absence of recognizable object content. Subsequent studies further applied the same evaluation strategy to biomedical image datasets~\citep{dhar2021evaluation} and astronomical image classification~\citep{dhar2022systematic}, demonstrating that CNNs continued to exploit contextual and acquisition-related information across fundamentally different application domains.

Most recently, this methodology has been applied to increasingly realistic settings: synthetically generated image datasets~\citep{martinez2026detection} and cancer pathology images~\citep{okonoda2026unmasking} were shown to permit above-chance classification from background-only regions, establishing background-based evaluation against random chance as a principled, quantitative framework for measuring contextual reliance. 


Several studies have suggested that CNNs trained for agricultural image classification rely on contextual information beyond agriculturally meaningful visual characteristics, though the evidence for this has so far been indirect. Mohanty et al.~\citep{mohanty2016using} showed that disease classifiers trained on laboratory images generalized poorly to field images, a performance drop consistent with reliance on acquisition conditions rather than disease symptoms alone, though the study did not isolate background content directly. Subsequent work used activation and saliency visualization to report that CNN predictions were influenced by healthy plant tissue, crop identity, and surrounding vegetation rather than disease-specific regions~\citep{atabay2017,ferentinos2018deep,toda2019convolutional,lee2020new}. 

Together, these studies indicate that contextual information can become correlated with agricultural class labels through imaging conditions and data collection practices. However, the evidence is largely limited to individual datasets and qualitative interpretation using visualization techniques, making it difficult to assess how prevalent contextual reliance is across agricultural image classification. To our knowledge, no prior work has systematically evaluated background-based classification across diverse agricultural datasets while accounting for differences in class cardinality, acquisition conditions, and dataset characteristics within a unified evaluation framework. This study addresses that gap through a comprehensive evaluation spanning eight agricultural benchmark datasets and four widely used CNN architectures.

\section{Methodology}
\label{method}

We designed a controlled evaluation framework to quantify the extent to which CNNs trained for agricultural and plant-related image classification rely on contextual information. The framework compares model performance on original test images with performance on small, background-dominated regions extracted from the same images. If a model primarily relies on object-specific characteristics, such as disease symptoms, plant morphology, or weed structure, predictions from these restricted regions should not consistently exceed dataset-specific random chance. Conversely, above-chance classification from regions containing limited information about the primary object indicates that contextual or acquisition-specific information contributes to the model predictions.

\begin{figure}
\centering
\includegraphics[width=0.90\linewidth]{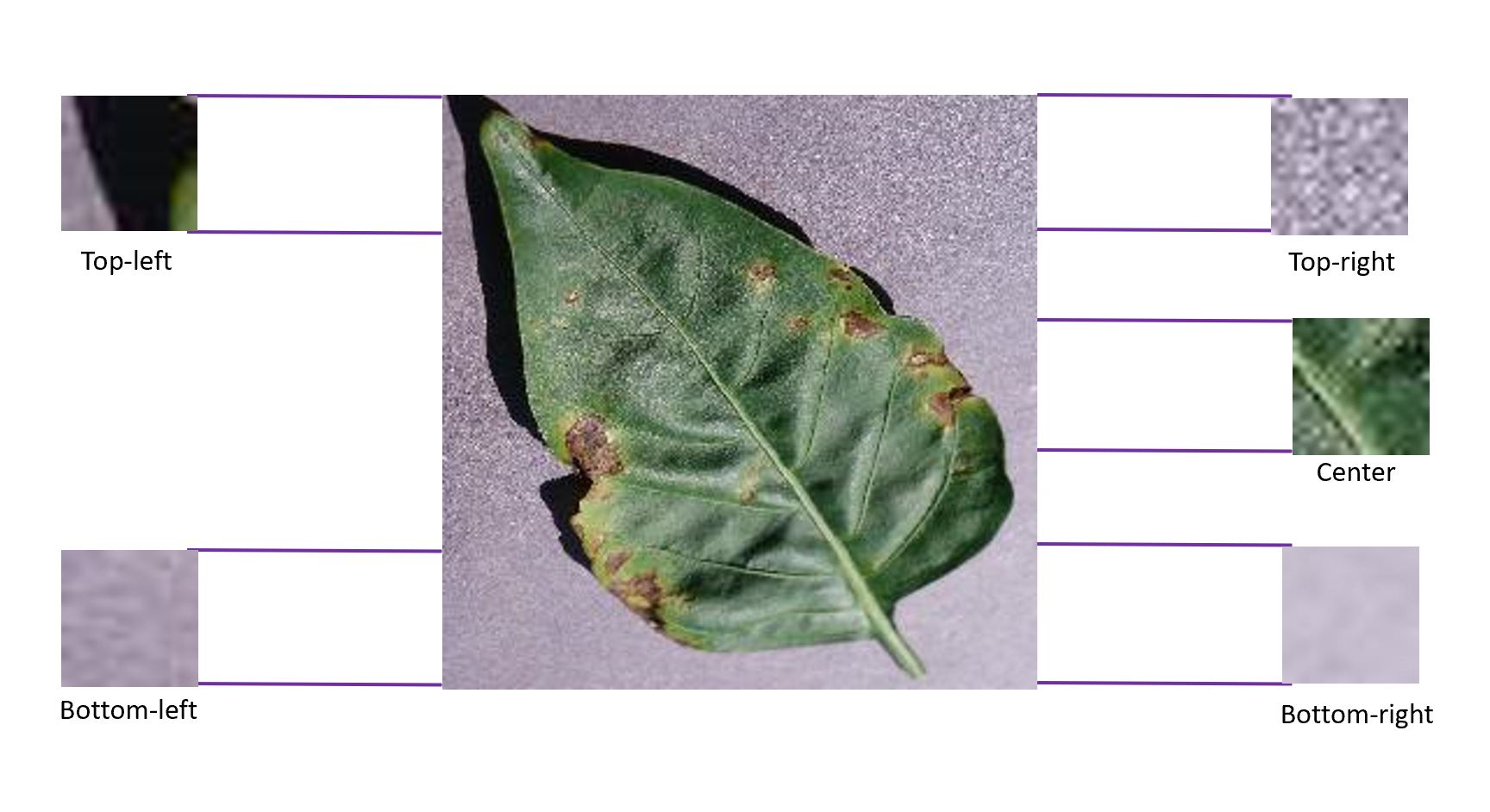}
\caption{\label{fig:extraction} Illustration of the patch extraction procedure using an image from PlantVillage. Five $20\times20$ pixel regions are extracted from the top-left, top-right, center, bottom-left, and bottom-right locations. Each patch retains the label of the original image and is evaluated using a CNN trained on full-content images.}
\end{figure}

We evaluate five spatial locations separately because contextual information may not be distributed uniformly across an image. Corner patches frequently capture elements such as soil, sky, surrounding vegetation, imaging surfaces, or acquisition artifacts. The center patch provides a less restrictive condition because it may contain part of the foreground object. We therefore refer to the resulting inputs collectively as \textit{background-dominated patches}, rather than assuming that every extracted patch is entirely free of object-related content. Figure~\ref{fig:extraction} summarizes the extraction procedure.

The framework was applied uniformly across eight publicly available benchmark image datasets using four CNN architectures. We evaluate patch performance relative to the random-chance baseline of each dataset, enabling comparison across classification tasks with substantially different numbers of classes. The following sections describe the datasets, architectures and training procedure, patch generation method, and evaluation protocol.

\begin{figure}
\centering
\includegraphics[width=0.90\linewidth]{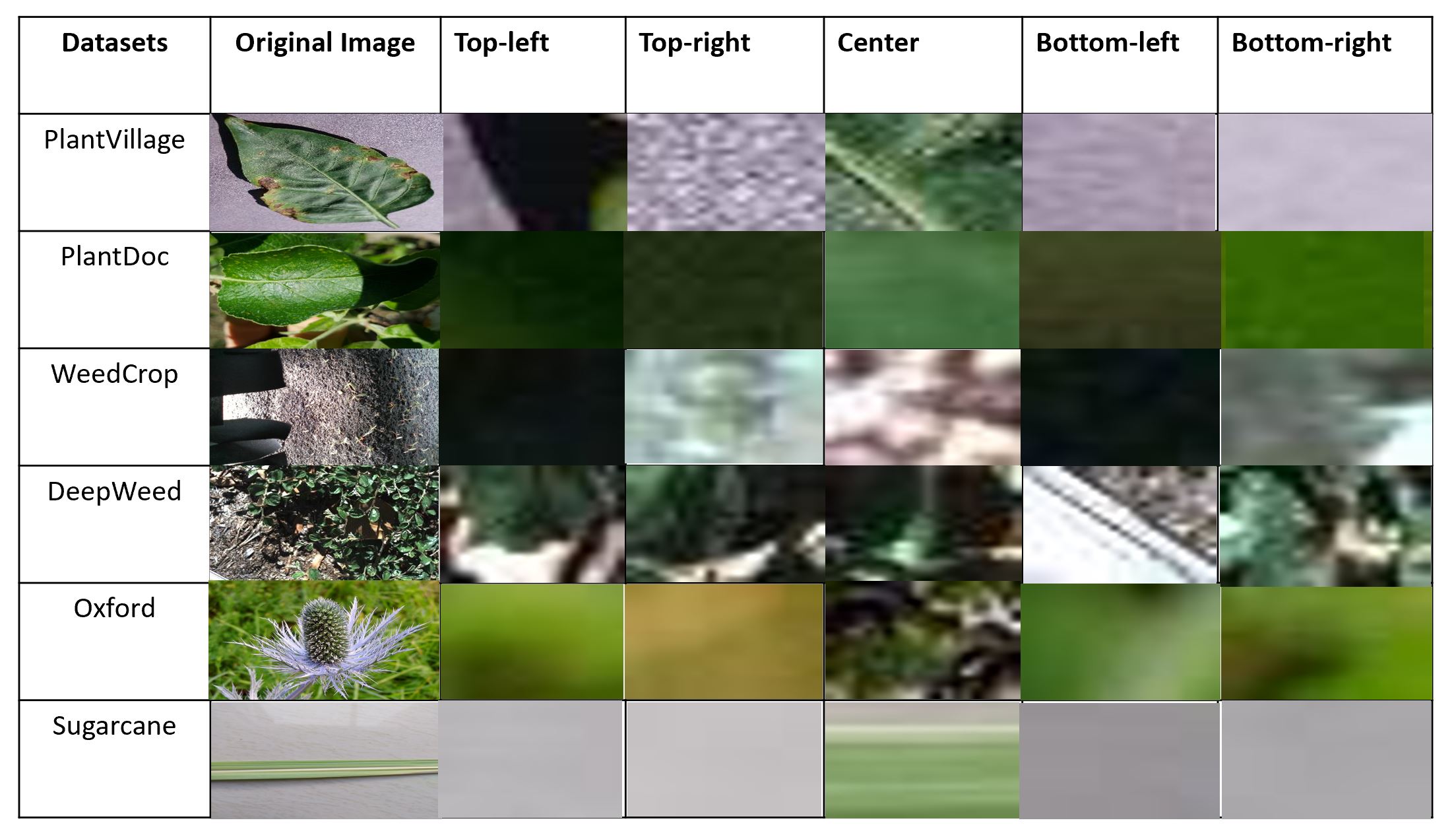}
\caption{\label{fig:diff_dataset} Examples of original images and corresponding $20\times20$ patches extracted from five spatial locations across six datasets. The patches retain background textures, acquisition characteristics, and, in some cases, limited foreground content.}
\end{figure}

\subsection{Datasets}
\label{subsec:dataset}

We evaluated contextual reliance using eight publicly available image datasets covering plant disease recognition, weed identification, crop classification, flower recognition, and agricultural spraying decisions. The datasets vary substantially in size, number of classes, image acquisition procedure, and visual complexity. They include images captured under controlled laboratory conditions, natural field environments, ground-based imaging systems, and aerial platforms. This heterogeneity allows us to examine whether contextual reliance is limited to particular acquisition settings or persists across different agricultural and plant-related classification tasks.

Seven datasets represent agricultural applications, while Oxford 102 Flower provides a plant-related fine-grained classification benchmark outside direct agricultural decision-making. Including datasets with different class cardinalities is particularly important for this study because raw patch accuracy cannot be compared directly across datasets. We therefore evaluate patch performance relative to a dataset-specific random-chance baseline, as described in Section~\ref{subsec:evaluation}. Table~\ref{tab:datasets} summarizes the datasets used in the experiments.

\begin{table*}[t]
\centering
\caption{Summary of the agricultural and plant-related image datasets used in the evaluation. Image counts correspond to the dataset versions used in the experiments.}
\label{tab:datasets}
\resizebox{\textwidth}{!}{%
\begin{tabular}{l|l| r| r| l}
\hline
\textbf{Dataset} & \textbf{Task} & \textbf{\# Images} &
\textbf{\# Classes} & \textbf{Acquisition setting} \\
\hline
PlantVillage & Disease classification & 54,000 & 15 & Controlled \\
PlantDoc &  Disease classification & 2,569 & 27 & Field \\
DeepWeeds &  Species classification & 17,509 & 9 & Field \\
Cassava &  Disease classification & 21,397 & 5 & Field \\
WeedCrop & Crop/weed classification & 2,822 & 2 & Field \\
Oxford 102  & Species classification & 8,189 & 102 & Natural/uncontrolled \\
Sugarcane & Disease classification & 3,000 & 11 & Field \\
OpenSprayer & Weed classification & 10,000 & 2 & Aerial/field \\
\hline
\end{tabular}
}
\end{table*}

\paragraph{PlantVillage} ~\citep{plantvillage} contains images of healthy and diseased leaves collected under controlled conditions and placed against relatively uniform backgrounds. Its standardized acquisition setting makes it possible to examine whether visually consistent backgrounds or imaging surfaces become associated with particular classes.

\paragraph{PlantDoc}
~\citep{plantdoc} contains plant disease images collected under natural conditions, with substantial variation in illumination, viewpoint, occlusion, surrounding vegetation, and background composition. It therefore provides a more visually complex disease-recognition setting than PlantVillage.

\paragraph{DeepWeeds}~\citep{DeepWeeds2019} contains images of invasive weed species collected in field environments in northern Australia. The dataset includes variation in plant appearance, vegetation density, soil, illumination, and acquisition location.

\paragraph{Cassava}~\citep{cassava} dataset contains field images representing cassava disease categories and healthy plants. Images frequently include contextual elements such as soil, surrounding plants, human hands, and heterogeneous field backgrounds.

\paragraph{WeedCrop}~\citep{weedCrop} contains RGB field images used for crop and weed recognition. Repeated crop arrangements, soil characteristics, vegetation structure, and imaging perspective may provide contextual information correlated with the target labels.

\paragraph{Sugarcane}~\citep{sugarcane} dataset contains healthy and diseased sugarcane images collected in field conditions. The images contain variation in leaf appearance, illumination, background vegetation, and acquisition conditions.

\paragraph{OpenSprayer}~\citep{openSprayer} contains agricultural images captured in operational field settings using an aerial or spraying platform. Its viewpoint and acquisition process differ from the ground-level plant and leaf datasets, allowing contextual reliance to be evaluated under an additional imaging condition.

\paragraph{Oxford 102}~\citep{oxford102}  contains 8,189 natural images from 102 flower categories, with substantial variation in scale, pose, illumination, and background. We include it as a plant-related fine-grained recognition dataset outside direct agricultural diagnosis or management.

\subsection{CNN Architectures and Training}
\label{subsec:arch}

We evaluated four widely used CNN architectures: ResNet50V2~\citep{he2016identity}, DenseNet121~\citep{huang2017densely}, InceptionV3~\citep{szegedy2016rethinking}, and VGG16~\citep{simonyan2015very}. These architectures represent different feature-extraction and connectivity designs, including residual connections, dense connections, multi-scale convolutional processing, and sequential convolutional blocks. Their inclusion allows us to examine whether contextual reliance is associated with a particular architectural design or occurs across substantially different CNN families.

All networks were initialized using ImageNet-pretrained weights and fine-tuned separately for each dataset. The classification layer was replaced to match the number of classes in the corresponding dataset. ResNet50V2, DenseNet121, and VGG16 received inputs resized to $224\times224$ pixels, while InceptionV3 received inputs resized to $299\times299$ pixels. Architecture-specific preprocessing functions were applied to both original images and extracted patches.

Models were trained using the Adam optimizer~\citep{kingma2014adam} with a learning rate of $0.001$ and a batch size of 32. Training continued for a maximum of 10 epochs, with early stopping based on
validation loss using a patience of 3 epochs and restoration of the
best-performing weights. Data augmentation was applied only to the training images and included random rotation, horizontal flipping, and brightness adjustment. No augmentation was applied during evaluation.

For each dataset, we used a 70/10/20  train/validation/test split. The validation split was used exclusively for early stopping; all reported results are computed on the held-out test split. The same split was retained across architectures so that differences in performance could be attributed to model behavior rather than variation in the evaluated samples. Each architecture was trained only on full-content images. The trained model was subsequently evaluated on the original test set and the five patch-based versions of that test set.

\subsection{Background-Dominated Patch Generation}
\label{subsec:bg_patch}

For every image in the test split, we extracted five square patches of size $20\times20$ pixels from the top-left, top-right, center, bottom-left, and bottom-right regions. 
The corner patches were aligned with the corresponding image boundaries, while the center patch was positioned around the spatial center of the image. The same extraction rule was used for all images and datasets.

Each extracted patch inherited the class label of its source image. Consequently, every original test set produced five patch-based test sets with identical sample counts and class distributions. For evaluation, the patches were resized to the input resolution of the corresponding CNN (224×224 for ResNet50V2, DenseNet121, VGG16; 299×299 for InceptionV3) using the interpolation setting implemented in the
Keras data-loading pipeline. Although resizing magnifies the low-resolution patch, it does not introduce new semantic information; it only produces an input compatible with the architecture.

The extraction procedure was designed to retain contextual and acquisition-specific characteristics while substantially reducing the amount of object-related information available to the classifier. Corner patches commonly contained soil, vegetation clutter, sky, shadows, uniform laboratory backgrounds, image borders, or acquisition artifacts. The center patch was retained as a separate spatial condition because it may include part of the primary object and therefore provides a less restrictive test of spatial dependence.

No augmentation was applied to the extracted patches. The models were evaluated on the patches without retraining or adaptation. Thus, above-chance patch performance reflects information learned during training on the original full-content images rather than patterns learned directly from the cropped test inputs.

Figure~\ref{fig:extraction} illustrates the five extraction locations, and Figure~\ref{fig:diff_dataset} presents examples from datasets with different acquisition settings.

\subsection{Evaluation Protocol}
\label{subsec:evaluation}

For each dataset--architecture combination, we evaluated the trained CNN on the original test set and on the five patch-based test sets corresponding to the top-left, top-right, center, bottom-left, and bottom-right image regions. Each patch retained the class label of its source image, allowing model performance on the restricted image regions to be evaluated using the same classification task as the original images.

The original test-set accuracy represents the conventional classification performance of the model when the complete image is available. We then evaluated the same trained model on each of the five patch-based test sets without retraining or adaptation. Comparing performance across these conditions allowed us to determine how much classification ability remained when the available visual information was restricted primarily to contextual image regions.

\subsubsection{Comparison with Random Chance}
\label{subsubsec:random}
Because the datasets used in this study differ substantially in the number of classes, patch accuracy cannot be interpreted using a single common threshold. We therefore compared patch performance with the random-chance accuracy associated with each dataset. For dataset $d$ containing $K_d$ classes, the uniform random-chance accuracy is:

\begin{equation}
C_d = \frac{1}{K_d}.
\end{equation}

Patch accuracy near the random-chance baseline indicates that the restricted image region provides little useful information for predicting the original class label. In contrast, consistently above-chance classification indicates that information retained within the patch remains predictive of the target class, providing evidence of contextual reliance.

For each dataset, we summarize patch performance using the range of accuracies observed across architectures and spatial locations together with the mean patch accuracy. These values are interpreted relative to the dataset-specific random-chance baseline rather than as absolute accuracies, allowing comparisons across datasets with different class cardinalities.

\subsubsection{Contextual Bias Categorization}
\label{subsubsec:bias_categorization}

For each dataset, we summarize patch performance across the four CNN architectures and five spatial locations. Let $A_{dmp}$ denote the patch accuracy for dataset $d$, architecture $m$, and patch location $p$. The mean patch accuracy for dataset $d$ is defined as

\begin{equation}
\bar{A}_d =
\frac{1}{MP}
\sum_{m=1}^{M}
\sum_{p=1}^{P}
A_{dmp},
\end{equation}

where $M=4$ is the number of CNN architectures, $P=5$ is the number of patch locations. 

We characterize the strength of contextual bias using the mean margin above chance,

\begin{equation}
\Delta_d = 100(\bar{A}_d-C_d),
\end{equation}

where $\Delta_d$ is expressed in percentage points and $C_d$ is the random chance accuracy defined in Section~\ref{subsubsec:random}.

We use an absolute margin rather than a fixed multiple of chance because the evaluated datasets span chance baselines ranging from less than 1\% for Oxford-102 to 50\% for the binary WeedCrop and OpenSprayer datasets. A ratio-based criterion can assign a large score to a small absolute improvement in datasets with many classes. For example, increasing accuracy from 1\% to 2\% doubles chance performance while adding only one percentage point of predictive accuracy. The absolute margin therefore provides a more interpretable basis for comparison across datasets with different class cardinalities.

Datasets are categorized as exhibiting \textit{high contextual bias} when $\Delta_d \geq 10$ percentage points, \textit{moderate contextual bias} when $3 \leq \Delta_d < 10$ percentage points, and \textit{low contextual bias} when $\Delta_d < 3$ percentage points. We adopt these thresholds as
operational categories for summarizing the distribution of margins observed
in the present benchmark. They are used to organize and interpret the
results rather than as universal thresholds for contextual bias. The
categories provide an aggregate dataset-level summary, while architecture-
and location-specific results are retained and examined separately in
Section~\ref{results}.

\subsubsection{Class-Imbalance and Robustness Analysis}

The primary evaluation in this study is based on classification accuracy relative to the dataset-specific random-chance baseline. However, because several of the evaluated datasets exhibit class imbalance, accuracy alone may not fully reflect model behavior. We therefore performed additional analyses for datasets categorized as exhibiting high contextual bias to determine whether the observed patch-based performance remained after controlling for class imbalance.

First, we computed macro-averaged precision, recall, and F1 score for both the original and patch-based test sets. Unlike accuracy, macro-averaged metrics assign equal weight to every class regardless of its frequency, allowing us to determine whether above-chance patch performance extends across the complete label space rather than being driven primarily by majority classes.

Second, we evaluated each high-bias dataset using a class-balanced test subset. For every original and patch-based test condition, each class was randomly undersampled without replacement to match the number of samples in the smallest class, using a fixed random seed of 42. This produced a single balanced subset containing an equal number of samples from every class. Accuracy together with macro-averaged precision, recall, and F1 score were then computed on the balanced subset. Because a single undersampling draw was used rather than repeated resampling, the reported values correspond to one reproducible balanced evaluation.

Finally, to provide a reference for chance performance under the same evaluation metrics, we generated random-prediction baselines for the original and patch-based test conditions. Whereas Section~\ref{subsec:evaluation} introduces the analytical random-chance accuracy used in the primary evaluation, the empirical baseline reported here extends the comparison to the additional evaluation metrics considered in the robustness analysis. For a dataset containing $K_d$ classes, each sample was assigned a class label drawn uniformly at random from the $K_d$ possible classes. To reduce variability associated with a single random assignment, the procedure was repeated for 20 independent trials using a fixed random seed of 42. Accuracy together with macro-averaged precision, recall, and F1 score were computed for every trial, and the reported random baseline corresponds to the mean across the 20 trials.

Together, these analyses provide complementary evidence regarding the robustness of the observed contextual bias and address the two research questions introduced in Section~\ref{intro}. \textbf{RQ1} is addressed by determining whether CNNs achieve above-chance classification on background-dominated patches. \textbf{RQ2} is addressed by comparing the strength of this behavior across datasets and architectures and by evaluating whether the strongest observed effects persist under macro-averaged evaluation, balanced test conditions, and random-prediction baselines.

\section{Results and Discussion}
\label{results}

\subsection{Classification from Background-Dominated Regions}
\label{sec:rq1}

\begin{figure}
\begin{subfigure}{\linewidth}
  \includegraphics[width=.5\linewidth]{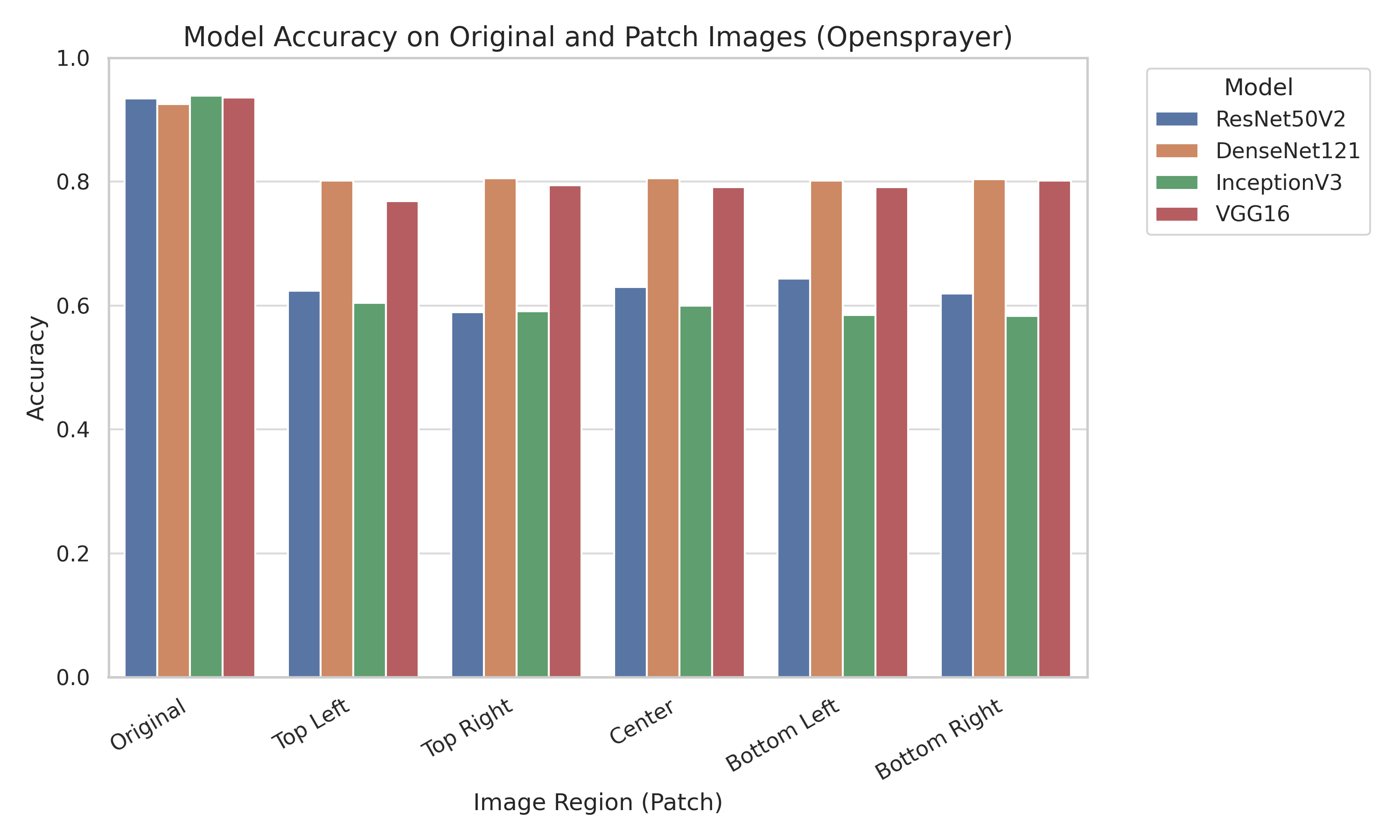}\hfill
  \includegraphics[width=.5\linewidth]{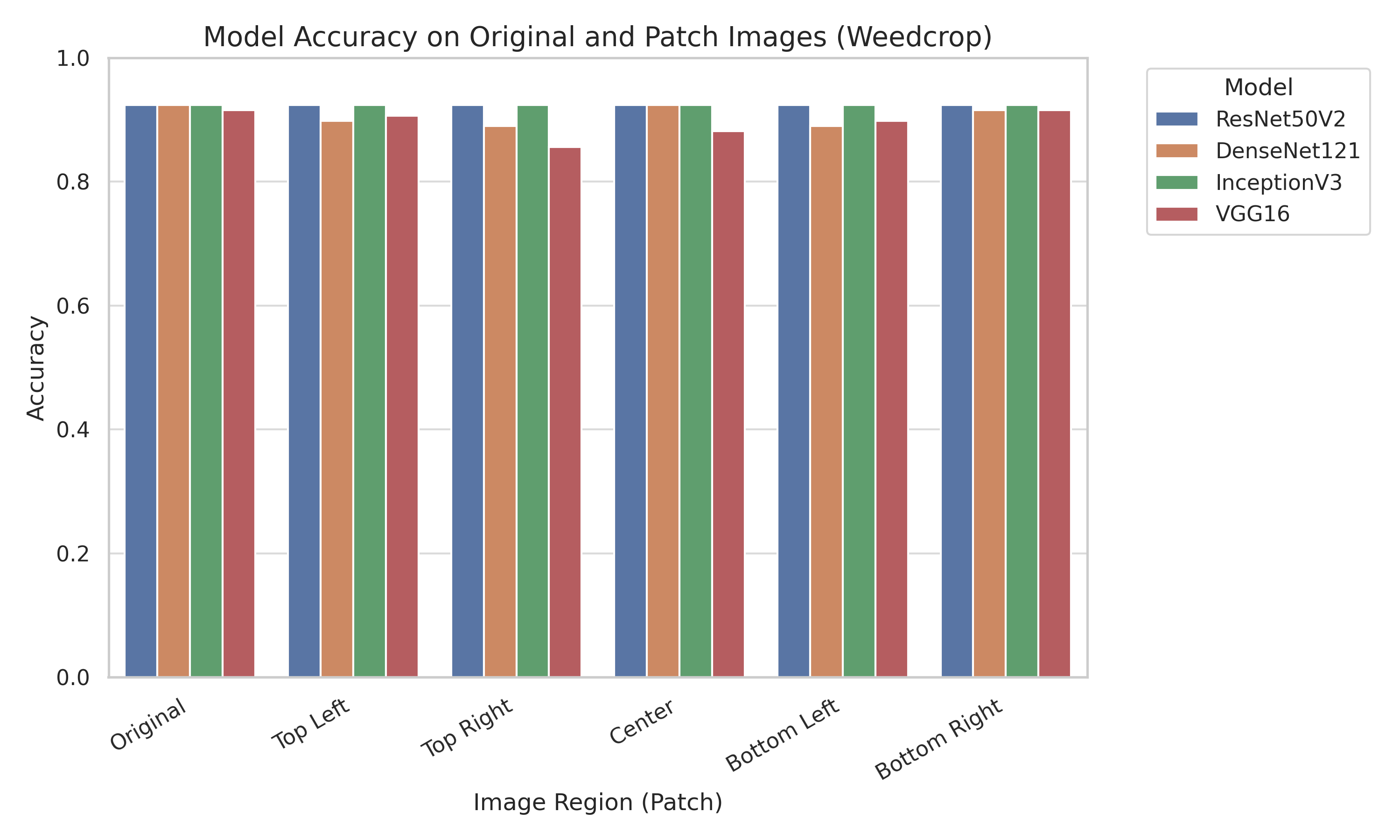}
  \end{subfigure}\par\medskip
 \begin{subfigure}{\linewidth}
  \includegraphics[width=.5\linewidth]{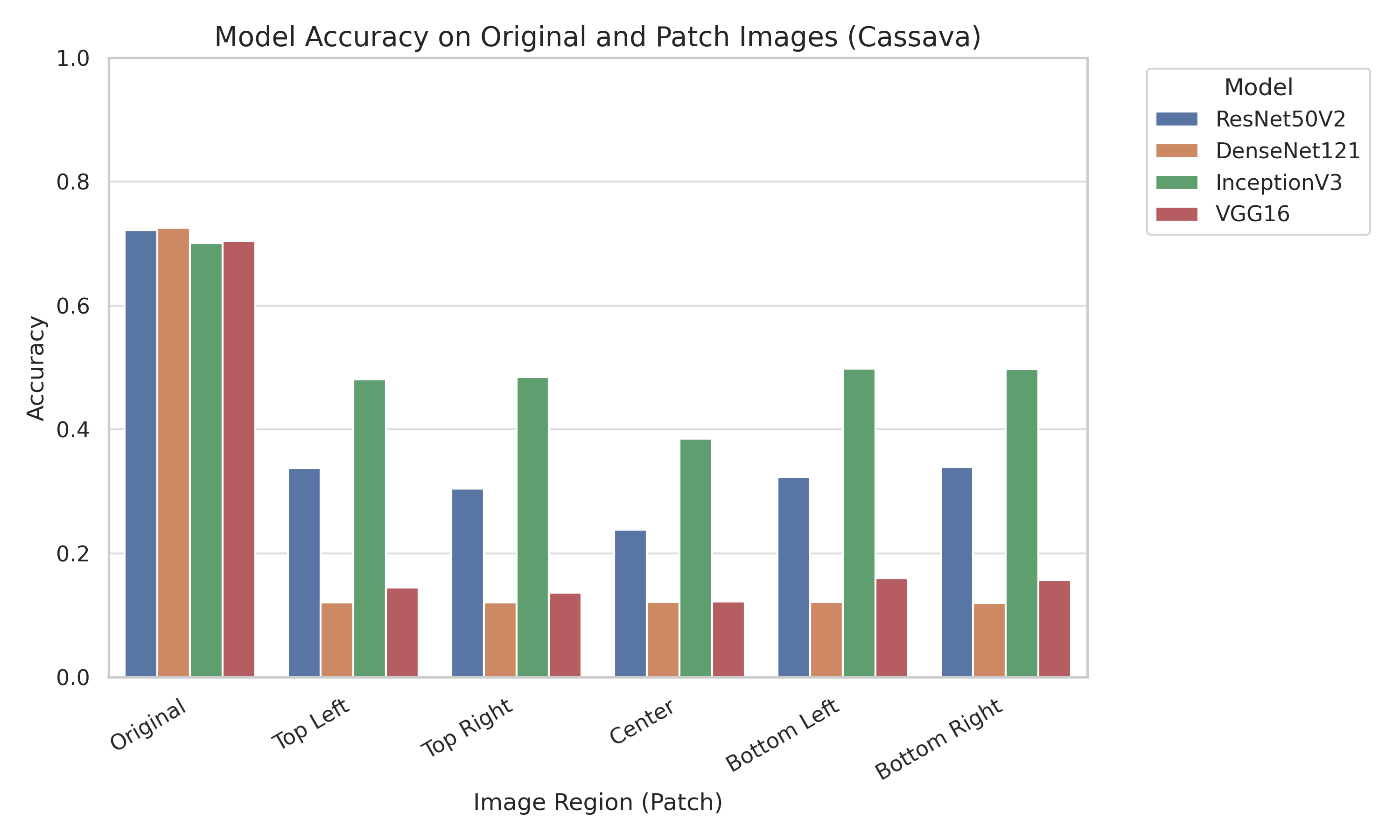}\hfill
  \includegraphics[width=.5\linewidth]{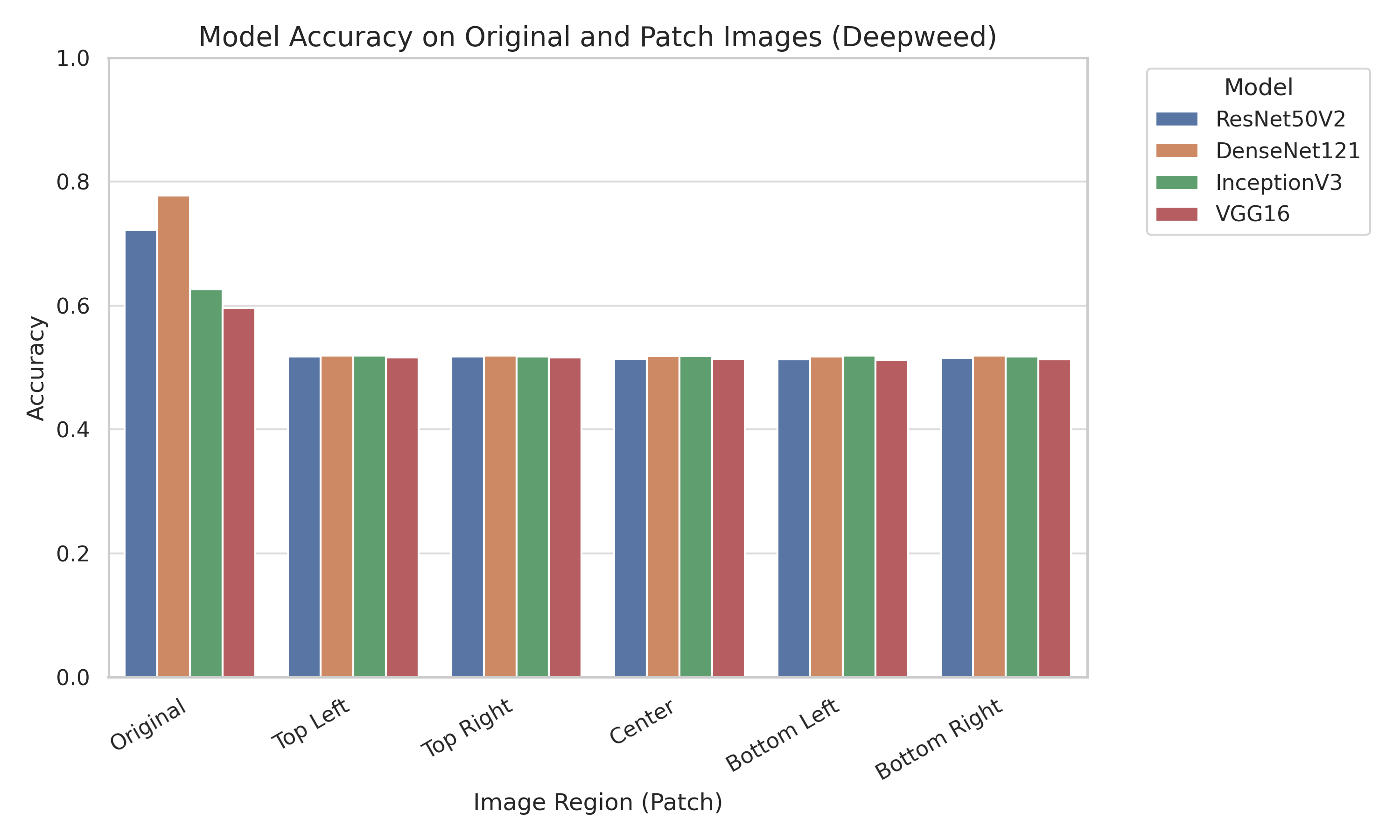}
  \end{subfigure}
  \caption{Original-image and patch-based classification accuracy for the four datasets exhibiting the strongest contextual reliance: OpenSprayer, WeedCrop, Cassava, and DeepWeeds (clockwise from top-left). Each panel reports accuracy for the original test set and for patches extracted from the top-left, top-right, center, bottom-left, and bottom-right image regions, across all four CNN architectures.}
  \label{fig:high_bias_accuracy}
\end{figure}

\begin{figure}
\begin{subfigure}{\linewidth}
  \includegraphics[width=.5\linewidth]{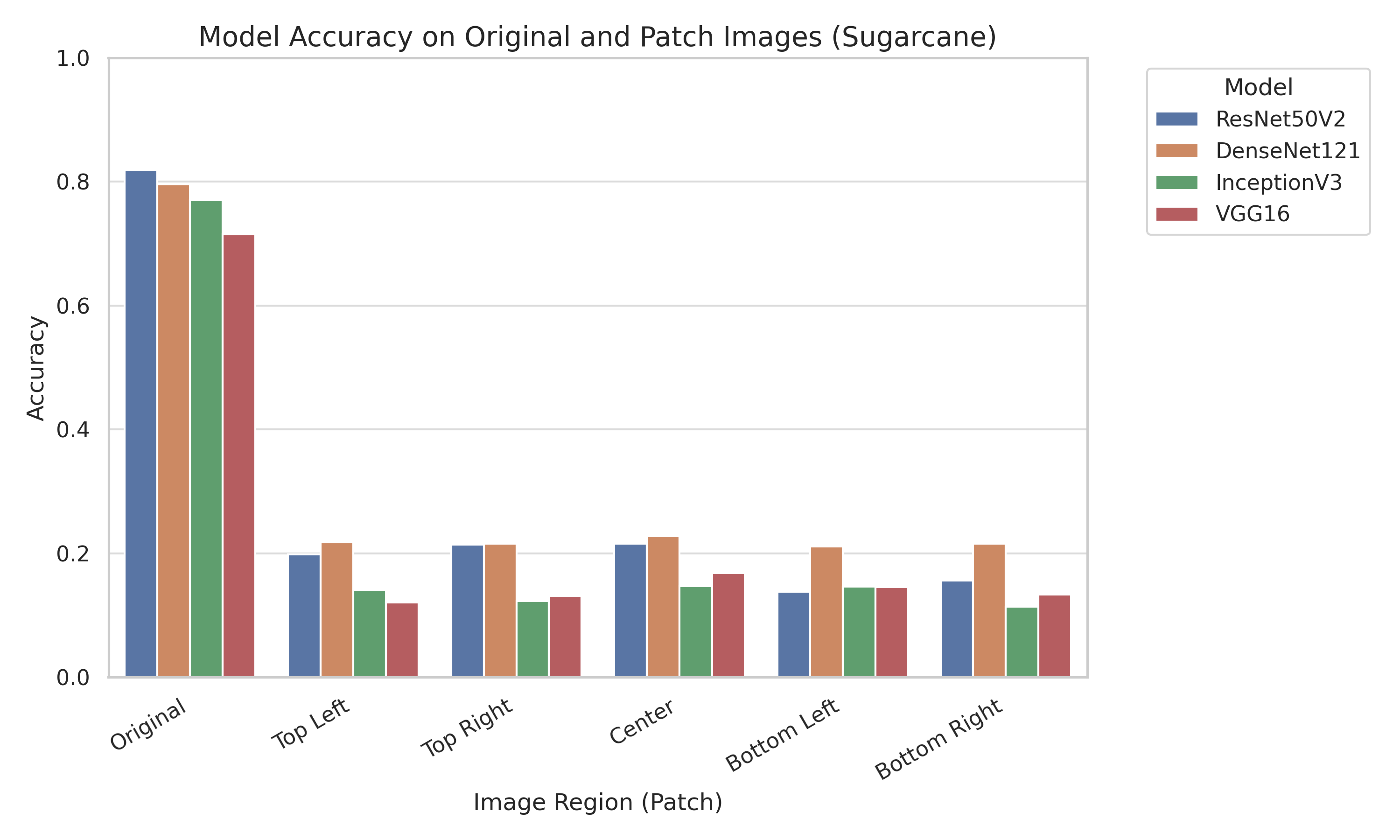}\hfill
  \includegraphics[width=.5\linewidth]{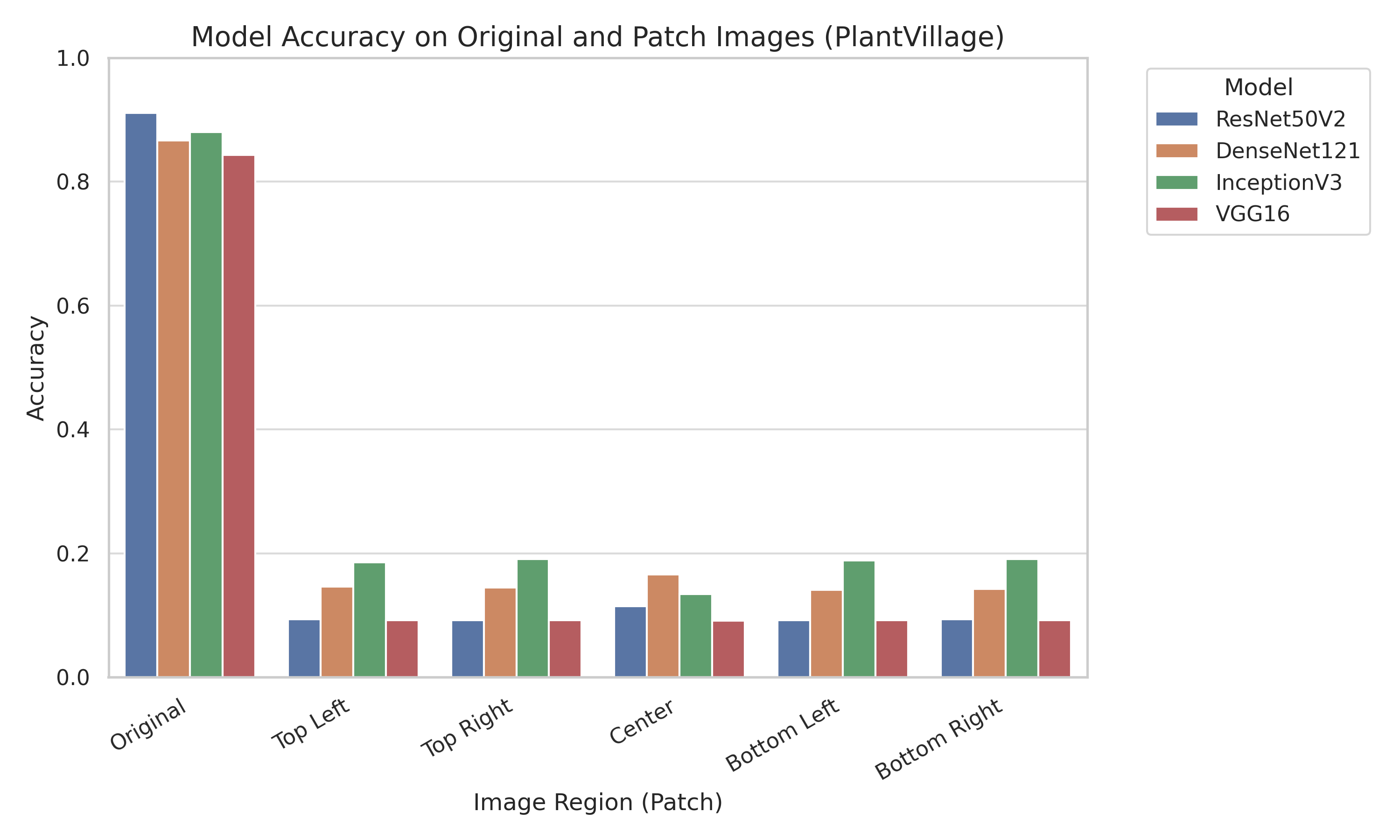}
  \end{subfigure}\par\medskip
 \begin{subfigure}{\linewidth}
  \includegraphics[width=.5\linewidth]{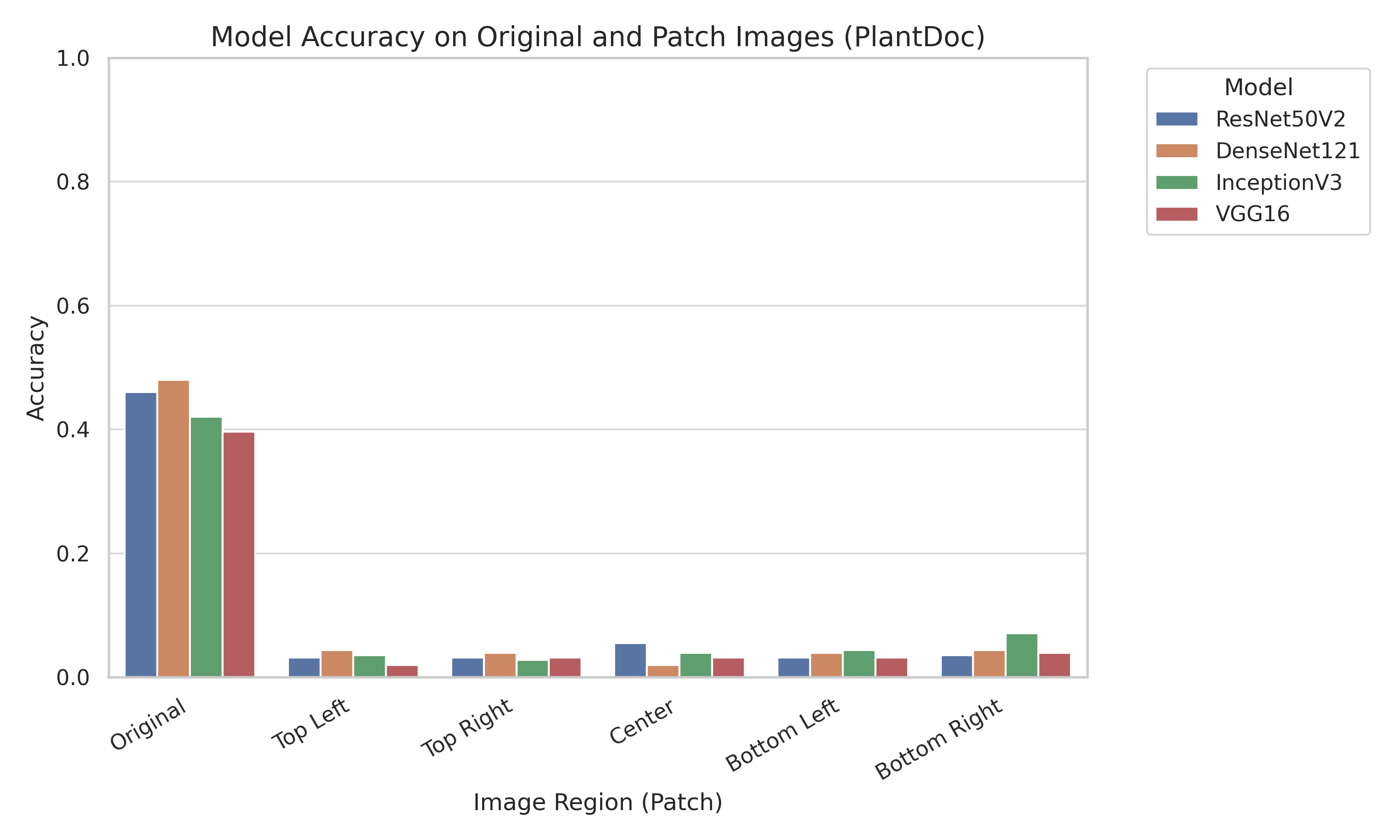}\hfill
  \includegraphics[width=.5\linewidth]{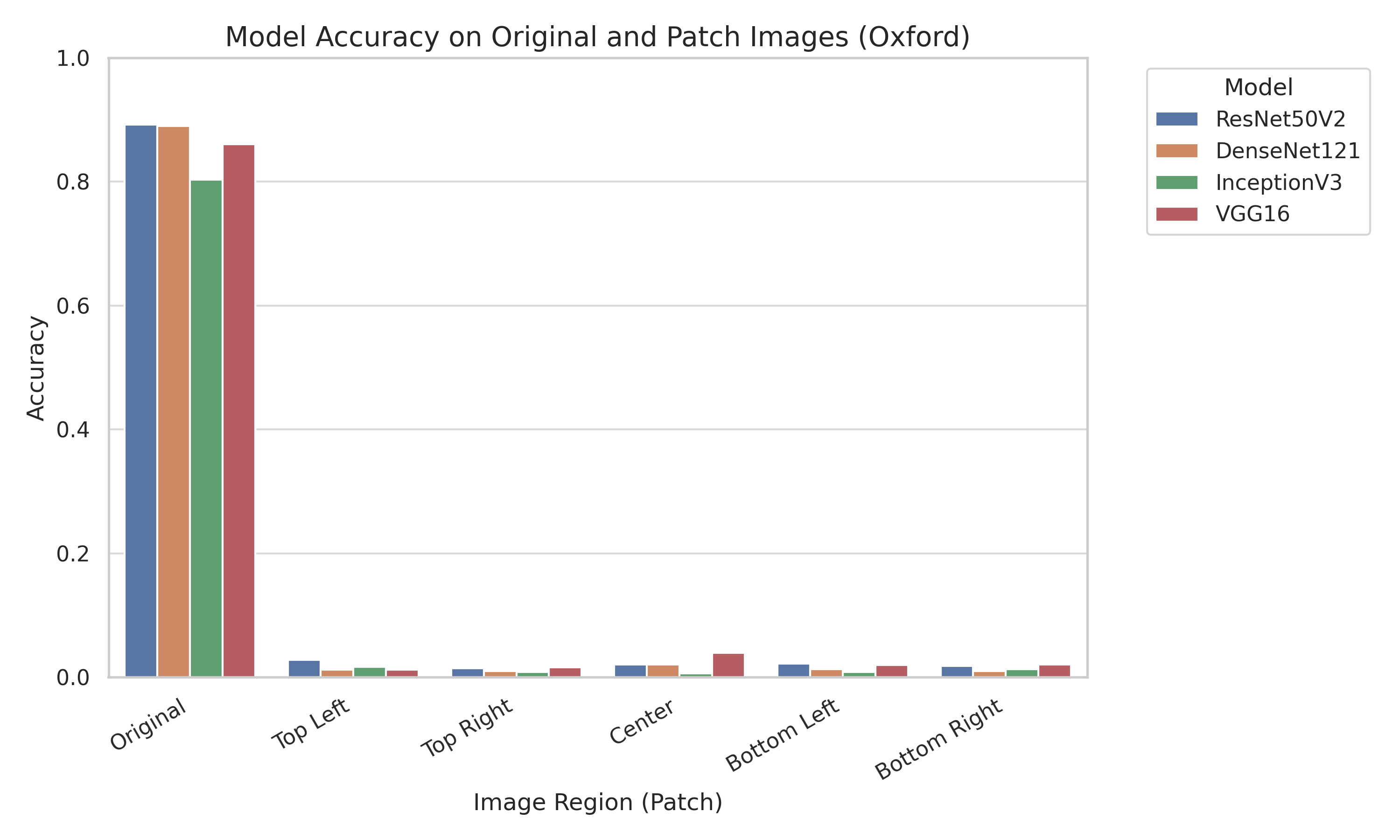}
  \end{subfigure}
  \caption{Original-image and patch-based classification accuracy for the four datasets showing weaker contextual reliance, evaluated across all four CNN architectures and five patch locations. Datasets are grouped by the contextual bias category assigned in Section~\ref{subsec:bias_category}: moderate (top) and low (bottom).}
\label{fig:low_bias_accuracy}
\end{figure}

To address \textbf{RQ1}, we first evaluate whether CNNs trained for agricultural image classification retain predictive capability when only background-dominated image regions are available during inference. For each of the eight datasets, four CNN architectures were trained using the original images and subsequently evaluated on the original test set together with five patch-based test sets extracted from different spatial locations. Figures~\ref{fig:high_bias_accuracy} and~\ref{fig:low_bias_accuracy} present the classification accuracies obtained for all datasets, while Table~\ref{tab:rq1_summary} summarizes the original-image accuracy, patch accuracy, and the corresponding dataset-specific random-chance baselines.

Substantial differences were observed across the eight evaluated datasets even though the patches contained little object-related visual information. WeedCrop exhibited the smallest reduction in performance, with the mean accuracy decreasing from 92.2\% on the original images to 91.1\% on the extracted patches. Similarly, OpenSprayer retained a mean patch accuracy of 70.2\% compared with an original-image accuracy of 93.4\%. DeepWeeds and Cassava showed larger reductions, from 68.1\% to 51.7\% and from 71.3\% to 33.6\%, respectively. Nevertheless, the patch accuracies for all four datasets remained substantially above their corresponding random-chance baselines.

The remaining datasets exhibited considerably lower patch performance. Sugarcane and PlantVillage achieved mean patch accuracies of 16.9\% and 12.9\%, exceeding their respective random-chance accuracies of 9.1\% and 6.7\%, but with considerably larger reductions relative to the original-image performance. In contrast, PlantDoc and Oxford-102 achieved mean patch accuracies of 3.7\% and 1.7\%, respectively, both remaining close to the expected random-chance performance for those datasets.

Although every dataset experienced a reduction in classification accuracy after replacing the original test images with background-dominated patches, the magnitude of that reduction varied considerably. Six of the eight evaluated datasets achieved mean patch accuracies above their dataset-specific random-chance baselines, whereas only PlantDoc and Oxford-102 remained effectively at chance performance. These results indicate that contextual information retained within the extracted patches contributed to CNN predictions for most datasets, although the strength of this behavior differed substantially across datasets. The following section quantifies these differences using the proposed contextual bias characterization framework.

\begin{table}
\centering
\caption{Summary of original-image and patch-based classification performance across the eight evaluated datasets. For each dataset, the table reports the number of classes, dataset-specific random-chance accuracy, mean original-image accuracy, the range and mean of patch accuracy across four CNN architectures and five spatial locations, the resulting margin above chance, and the corresponding contextual bias category (High, Moderate, or Low), as defined in Section~\ref{subsubsec:bias_categorization}.}
\label{tab:rq1_summary}
\resizebox{0.9\textwidth}{!}{%
\begin{tabular}{l|c|c|c|c|c|c|c}
\hline
\textbf{Dataset} & \textbf{\# Classes} & \textbf{Random Chance} & \textbf{Mean Original} & \textbf{Patch Acc. Range} & \textbf{Mean Patch} & \textbf{Margin} & \textbf{Category}\\ \hline
WeedCrop & 2 & 50.0\% & 92.2\%& 85.6–92.4\% & 91.1\% &	41.1 pp & High\\
DeepWeed & 9 & 11.1\% & 68.1\% & 51.3–55.0\% & 51.7\% &	40.6 pp & High\\
OpenSprayer & 2 & 50.0\% & 93.4\% & 58.4–80.6\% & 70.2\% &20.2 pp & High\\
Cassava & 5 & 20.0\% & 71.3\% & 16.0–55.8\% & 33.6\% & 13.6 pp & High\\
Sugarcane & 11 & 9.1\% & 77.5\%& 11.4–22.8\% & 16.9\% &	7.8 pp & Moderate\\
PlantVillage & 15 & 6.7\% &	87.5\% & 9.1–19.1\% & 12.9\% &	6.2 pp & Moderate\\
Oxford-102 & 102 & 0.98\% &	86.2\% & 0.6–3.9\% & 1.7\% & 0.7 pp & Low\\
PlantDoc & 27 & 3.7\% &	44.0\% & 2.0–7.1\% & 3.7\% & 0.03 pp & Low\\
\hline
\end{tabular}%
}
\end{table}

\subsection{Dataset-level characterization of contextual bias}
\label{subsec:bias_category}
Section~\ref{sec:rq1} established that CNNs retained above-chance classification performance on background-dominated patches for six of the eight evaluated datasets. We now quantify the strength of this behavior using the contextual bias characterization framework introduced in Section~\ref{subsubsec:bias_categorization}. Table~\ref{tab:rq1_summary} summarizes the resulting margins above chance together with the corresponding contextual bias category assigned to each dataset.

Based on the mean margin above chance, four datasets were categorized as exhibiting high contextual bias. WeedCrop (41.1 pp) and DeepWeeds (40.6 pp) produced the largest margins, followed by OpenSprayer (20.2 pp) and Cassava (13.6 pp). Although all four datasets substantially exceeded their respective random-chance baselines, the magnitude of the effect differed considerably, indicating varying degrees of contextual reliance. 

Sugarcane and PlantVillage formed a second group characterized by moderate contextual bias, with margins of 7.8 and 6.2 percentage points, respectively. While both datasets consistently exceeded random chance, their margins remained well below those observed for the high-bias group. 

Oxford-102 and PlantDoc exhibited low contextual bias, with margins of only 0.7 and 0.03 percentage points above chance. For these datasets, classification performance on background-dominated patches approached the corresponding random-chance baseline, providing little evidence that contextual information alone was predictive of the original class labels. 

Although the proposed categorization summarizes each dataset using the mean patch accuracy across four CNN architectures and five patch locations, the underlying results exhibit varying degrees of consistency. For example, the patch accuracy ranges reported in Table~\ref{tab:rq1_summary} are comparatively narrow for DeepWeeds and WeedCrop, whereas Cassava and OpenSprayer exhibit substantially larger ranges, indicating greater variability across architectures and spatial locations. Consequently, the dataset-level margins reported in this section provide an aggregate measure of contextual bias rather than a complete description of its behavior under individual experimental conditions. Because the strongest contextual reliance was observed for the four high-bias datasets, the remaining analyses focus exclusively on these datasets to determine whether the observed effect persists under more rigorous evaluation.


\subsection{Class-Sensitive Evaluation of High-Bias Datasets}
\label{subsec:class_sensitive}

\begin{figure}
\centering
\includegraphics[trim=0.3cm 2.3cm 0.2cm 0.2cm, clip, width=\textwidth]{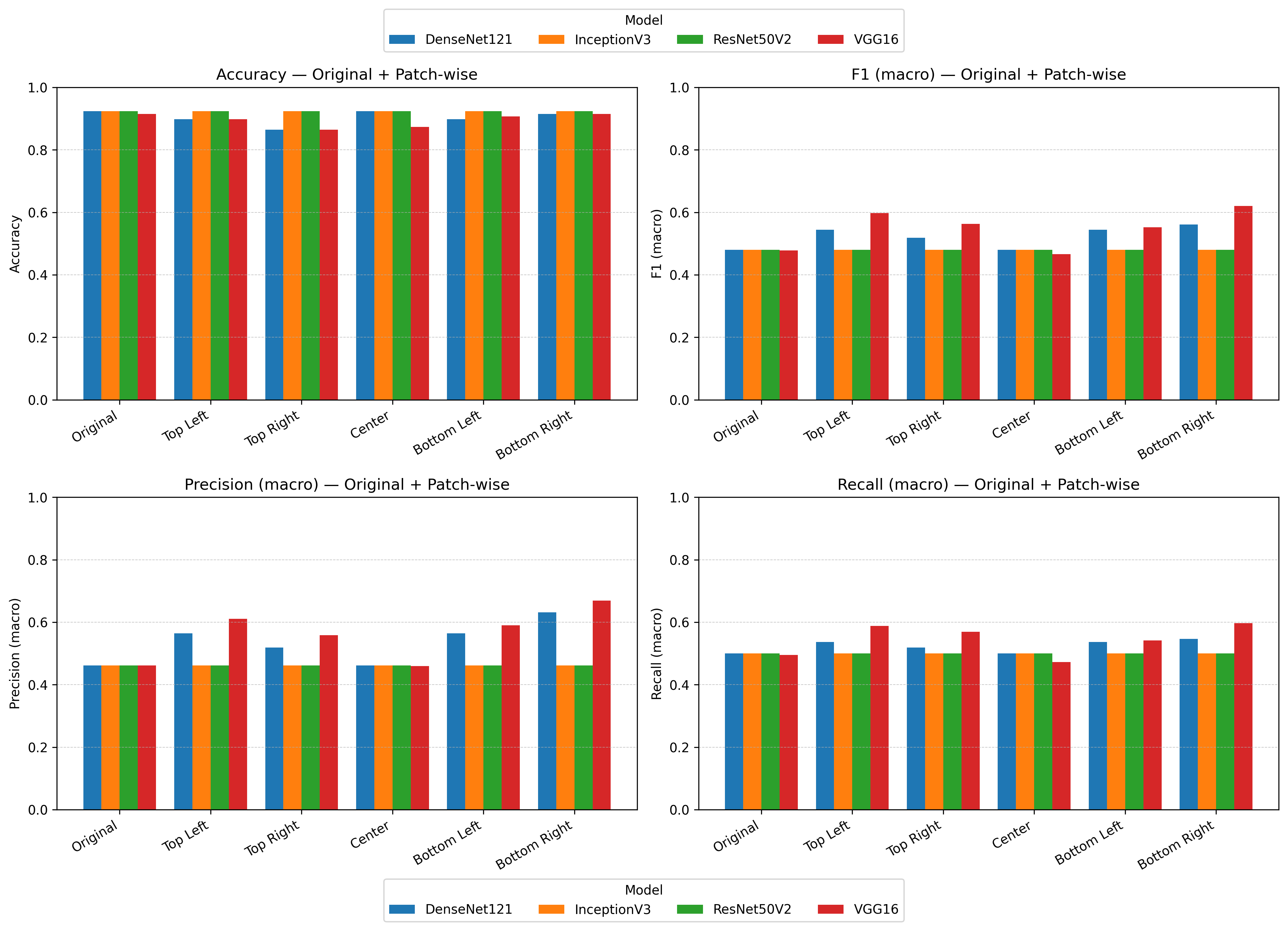}
\caption{\label{fig:weedcrop_macro}Performance of the four CNN architectures on the WeedCrop dataset. Classification accuracy (top left), macro-averaged F1 score (top right), macro-averaged precision (bottom left), and macro-averaged recall (bottom right) are reported for the original test images and the five background-dominated patch locations (top-left, top-right, center, bottom-left, and bottom-right).}
\end{figure}

\begin{figure}[ht]
\centering
\includegraphics[trim=0.3cm 2.3cm 0.2cm 0.2cm, clip, height=0.37\textheight, keepaspectratio]{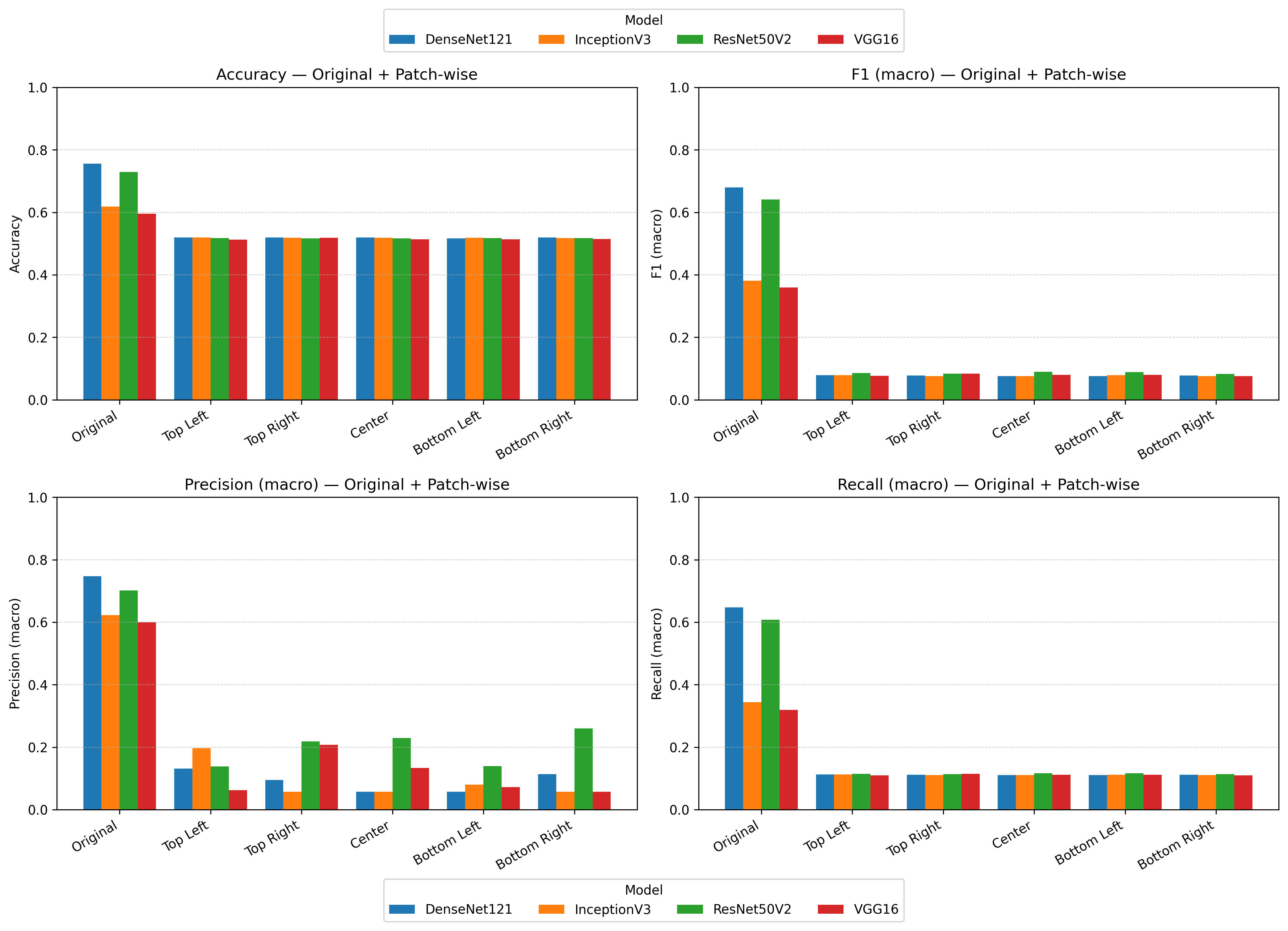}
\caption{\label{fig:deepweed_macro} Performance of the four CNN architectures on the DeepWeeds dataset. Classification accuracy (top left), macro-averaged F1 score (top right), macro-averaged precision (bottom left), and macro-averaged recall (bottom right) are reported for the original test images and the five background-dominated patch locations (top-left, top-right, center, bottom-left, and bottom-right).}
\end{figure}

\begin{figure}[ht]
\centering
\includegraphics[trim=0.3cm 2.3cm 0.2cm 0.2cm, clip, height=0.40\textheight, keepaspectratio]{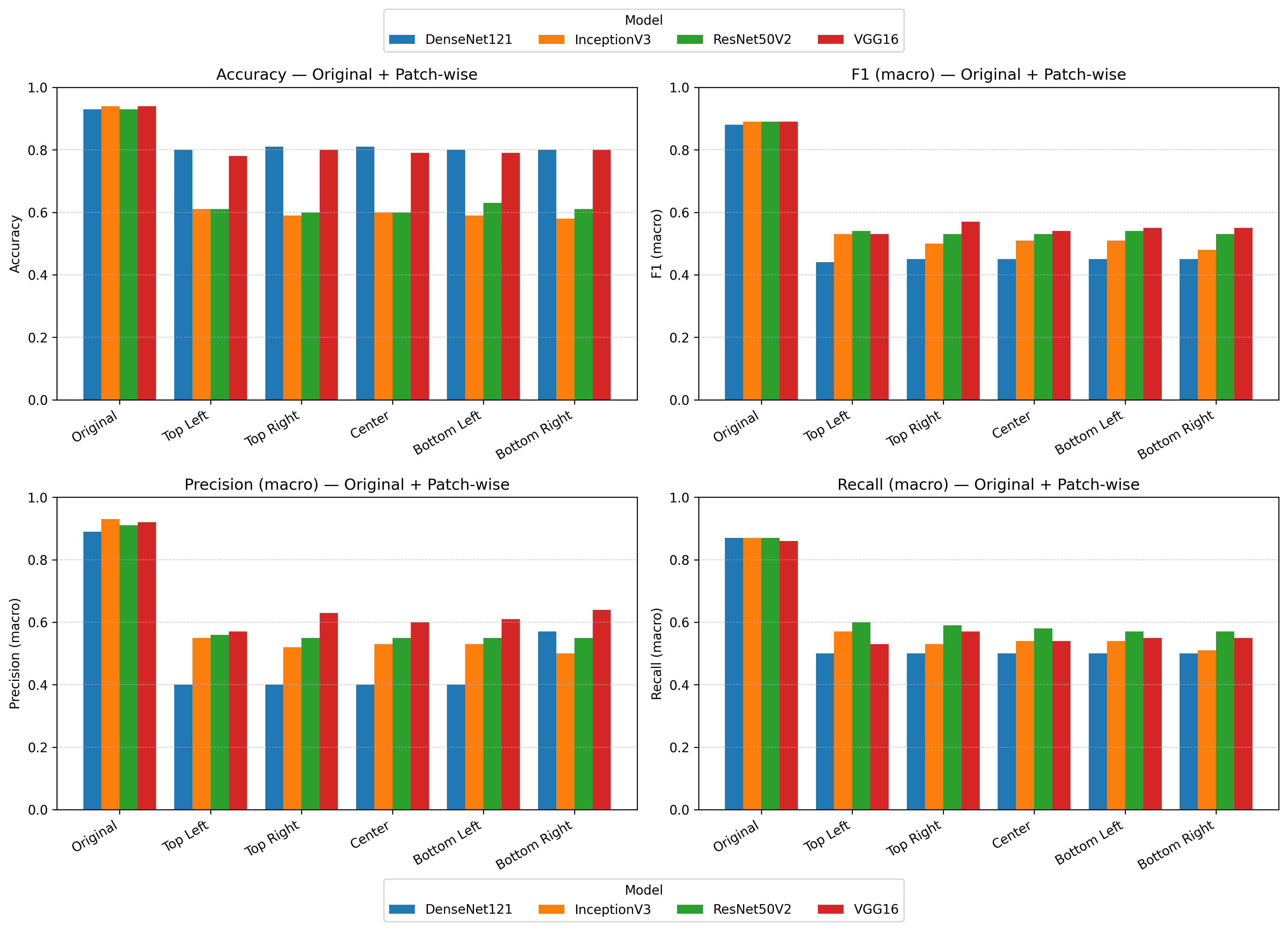}
\caption{\label{fig:opensprayer_macro} Performance of the four CNN architectures on the OpenSprayer dataset. Classification accuracy (top left), macro-averaged F1 score (top right), macro-averaged precision (bottom left), and macro-averaged recall (bottom right) are reported for the original test images and the five background-dominated patch locations (top-left, top-right, center, bottom-left, and bottom-right).}
\end{figure}

\begin{figure}[ht]
\centering
\includegraphics[trim=0.3cm 2.3cm 0.2cm 0.2cm, clip, height=0.40\textheight, keepaspectratio]{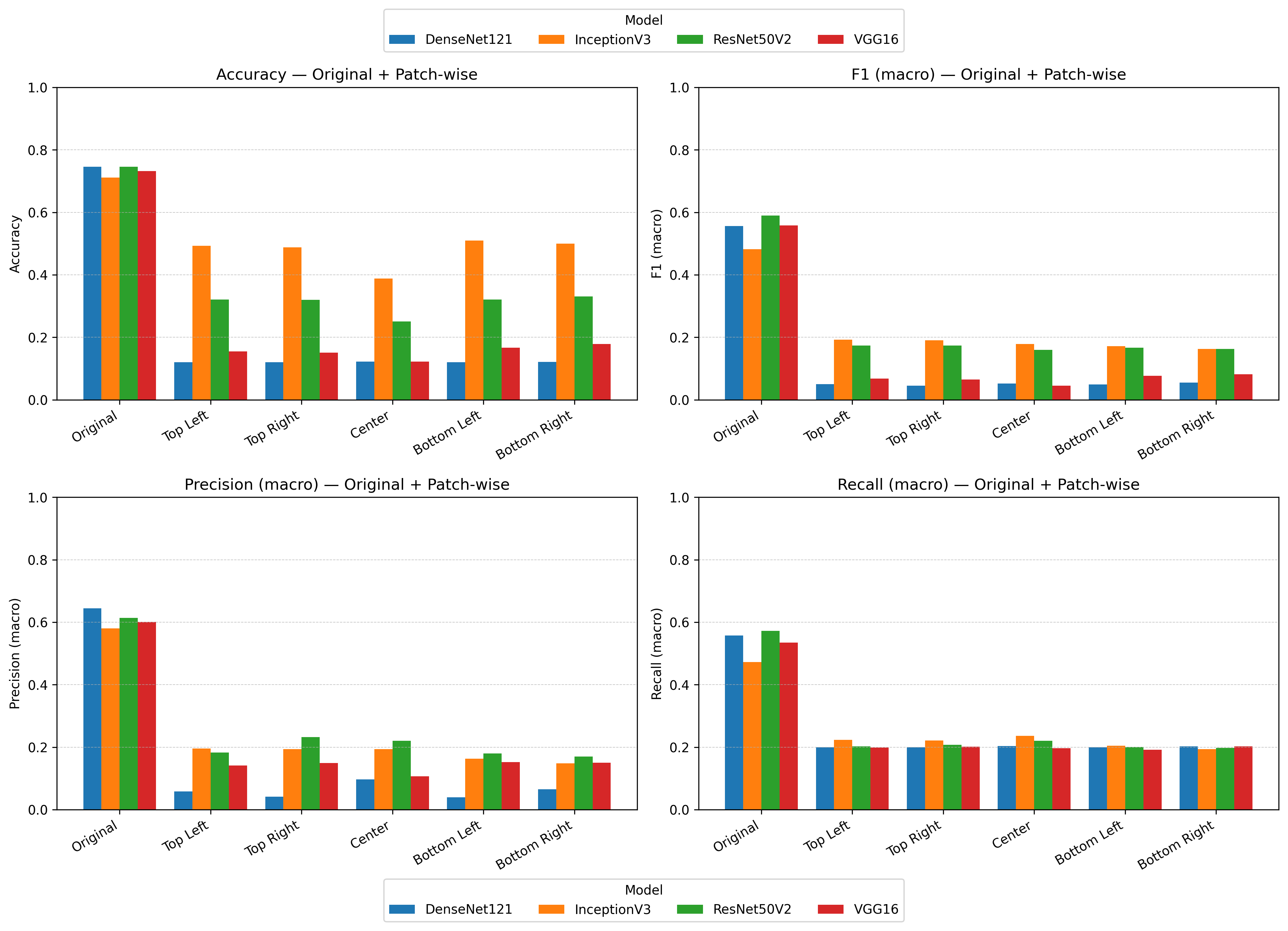}
\caption{\label{fig:cassava_macro} Performance of the four CNN architectures on the Cassava dataset. Classification accuracy (top left), macro-averaged F1 score (top right), macro-averaged precision (bottom left), and macro-averaged recall (bottom right) are reported for the original test images and the five background-dominated patch locations (top-left, top-right, center, bottom-left, and bottom-right).}
\end{figure}

The contextual bias categorization in Section~\ref{subsec:bias_category}
is based primarily on classification accuracy. However, accuracy can be
dominated by frequently occurring classes and may therefore overstate the
extent to which patch-based predictions generalize across the complete
label space. We consequently evaluated WeedCrop, DeepWeeds, OpenSprayer, and Cassava using macro-averaged precision, recall, and F1 score. To facilitate interpretation of these metrics, Table~\ref{tab:random_baseline_macro} reports the corresponding random-prediction baselines obtained from the evaluation procedure described in Section~\ref{subsec:evaluation}.
Figures~\ref{fig:weedcrop_macro}--\ref{fig:cassava_macro} present the
results for the original and patch-based test conditions.

\begin{table}[t]
\centering
\caption{Random-prediction baselines for the high-contextual-bias datasets. Values correspond to the mean of 20 independent random-prediction trials and provide reference performance for the macro-averaged evaluation metrics.}
\label{tab:random_baseline_macro}
\begin{tabular}{lcccc}
\hline
\textbf{Dataset} & \textbf{Accuracy \%} & \textbf{Precision \%} & \textbf{Recall \%} & \textbf{F1 \%} \\
\hline
WeedCrop     & 50.0 & 50.0 & 50.0 & 38.7 \\
DeepWeeds    & 11.1 & 11.1 & 11.2 & 09.0 \\
OpenSprayer  & 49.8 & 50.1 & 50.2 & 44.8 \\
Cassava      & 20.0 & 19.8 & 19.7 & 16.1 \\
\hline
\end{tabular}
\end{table}

The relationship between classification accuracy and the macro-averaged metrics differed substantially across the four high-bias datasets. Although WeedCrop retained patch accuracies close to 90\% across architectures and spatial locations, its macro F1 values ranged approximately from 45\% to 62\%. Importantly, these values remained consistently above the random-prediction baseline of 38.7\% (Table~\ref{tab:random_baseline_macro}), indicating that the observed patch performance reflected meaningful discrimination between the two classes rather than chance predictions alone.

A different pattern was observed for DeepWeeds. Despite maintaining patch accuracies of approximately 52\%, macro recall remained close to the random baseline (11.2\%), while macro F1 remained around the corresponding random baseline of 09.0\% across all patch conditions. Thus, the above-chance accuracy reported in Section~\ref{sec:rq1} did not translate into balanced discrimination across the nine classes, suggesting that predictive performance was concentrated in only a subset of the label space.

OpenSprayer showed the strongest agreement between accuracy and the class-sensitive metrics. Patch accuracies remained between approximately 60\% and 80\%, while macro precision, recall, and F1 consistently exceeded their corresponding random baselines (50.1\%, 50.2\%, and 44.8\%, respectively). This indicates that contextual information remained informative for both classes rather than merely increasing overall accuracy.

Cassava exhibited the greatest discrepancy between aggregate accuracy and class-sensitive evaluation. Although mean patch accuracy remained above chance, macro recall stayed close to the five-class random baseline (19.7\%), while macro F1 generally remained only slightly above its random baseline of 16.1\%. These results indicate that the contextual information retained within the patches provided substantially weaker and less uniformly distributed discriminative information across the five cassava disease classes than suggested by accuracy alone.

Overall, the macro-averaged evaluation refines the dataset-level characterization derived from classification accuracy by revealing how contextual information is distributed across the label space. While OpenSprayer retained comparatively strong performance across all evaluation metrics, WeedCrop, DeepWeeds, and Cassava exhibited substantially larger reductions in macro-averaged precision, recall, and F1 score than in overall accuracy, indicating that the contextual information retained in the extracted patches was not equally informative for all classes. Because these results are still evaluated on the original class distributions, the following section examines the same four datasets using class-balanced test subsets to determine the extent to which the observed behavior is influenced by class imbalance.

\subsection{Robustness under Class-Balanced Evaluation}
\label{subsec:balanced}

\begin{figure}
\centering
\includegraphics[trim=0.3cm 2.3cm 0.2cm 0.2cm, clip, height=0.40\textheight, keepaspectratio]{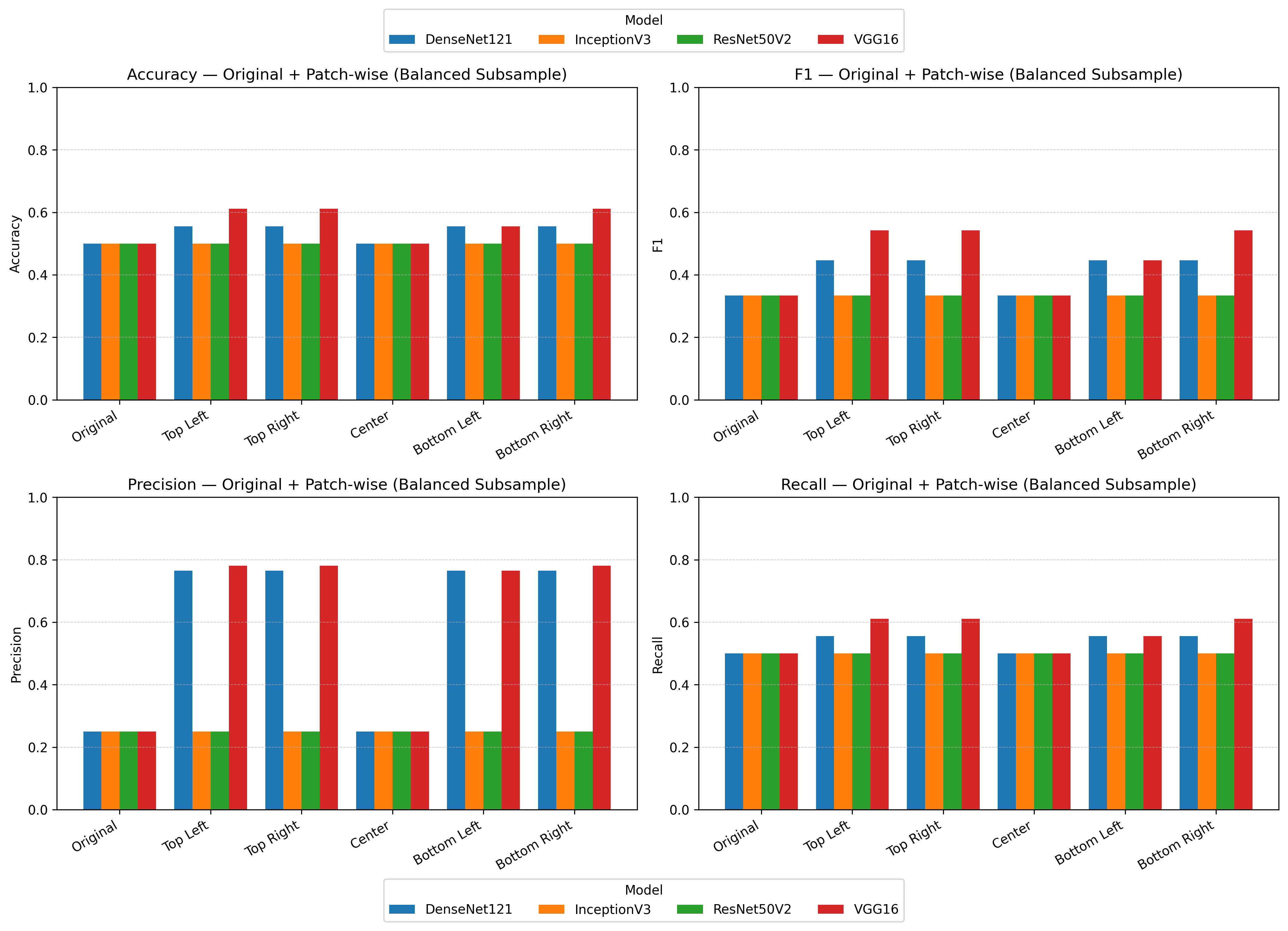}
\caption{\label{fig:weedcrop_bal}Performance of the four CNN architectures on the WeedCrop dataset. Classification accuracy (top left), macro-averaged F1 score (top right), macro-averaged precision (bottom left), and macro-averaged recall (bottom right) are reported for the original test images and the five background-dominated patch locations (top-left, top-right, center, bottom-left, and bottom-right).}
\end{figure}

\begin{figure}
\centering
\includegraphics[trim=0.3cm 2.3cm 0.2cm 0.2cm, clip, height=0.40\textheight, keepaspectratio]{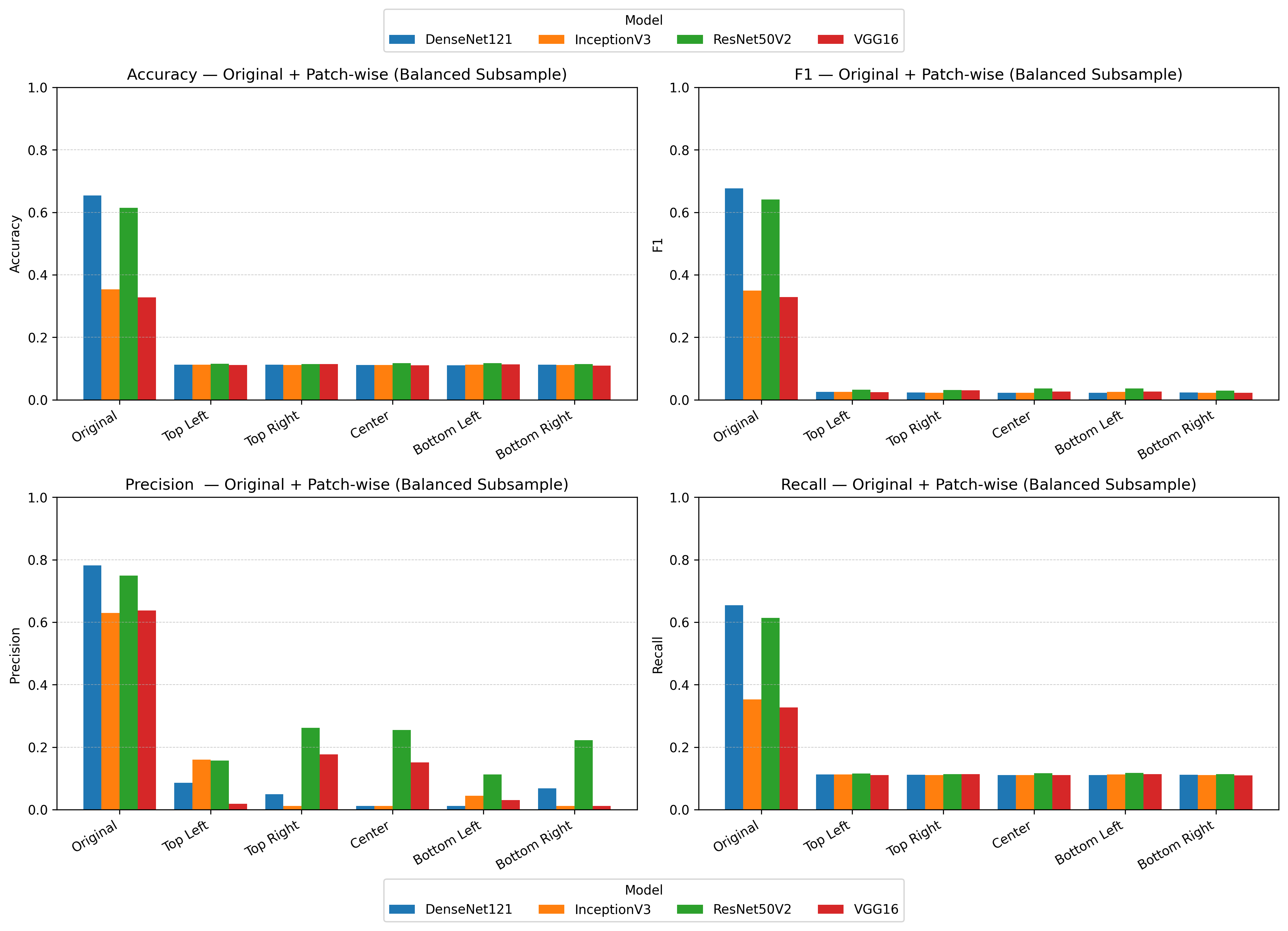}
\caption{\label{fig:deepweed_bal} Performance of the four CNN architectures on the DeepWeeds dataset. Classification accuracy (top left), macro-averaged F1 score (top right), macro-averaged precision (bottom left), and macro-averaged recall (bottom right) are reported for the original test images and the five background-dominated patch locations (top-left, top-right, center, bottom-left, and bottom-right).}
\end{figure}

\begin{figure}
\centering
\includegraphics[trim=0.3cm 2.3cm 0.2cm 0.2cm, clip,height=0.40\textheight, keepaspectratio]{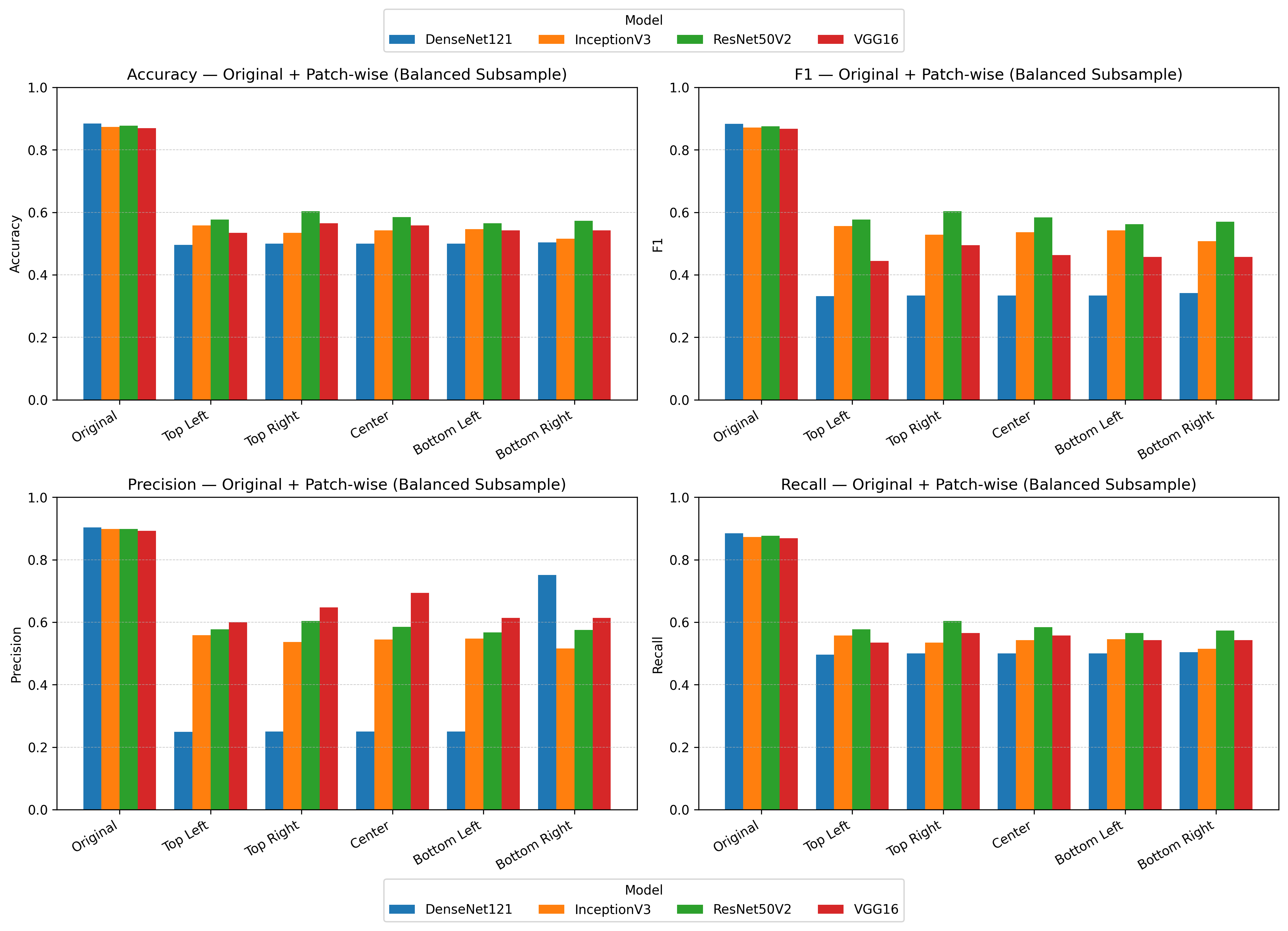}
\caption{\label{fig:opensprayer_bal} Performance of the four CNN architectures on the OpenSprayer dataset. Classification accuracy (top left), macro-averaged F1 score (top right), macro-averaged precision (bottom left), and macro-averaged recall (bottom right) are reported for the original test images and the five background-dominated patch locations (top-left, top-right, center, bottom-left, and bottom-right).}
\end{figure}

\begin{figure}
\centering
\includegraphics[trim=0.3cm 2.3cm 0.2cm 0.2cm, clip, height=0.40\textheight, keepaspectratio]{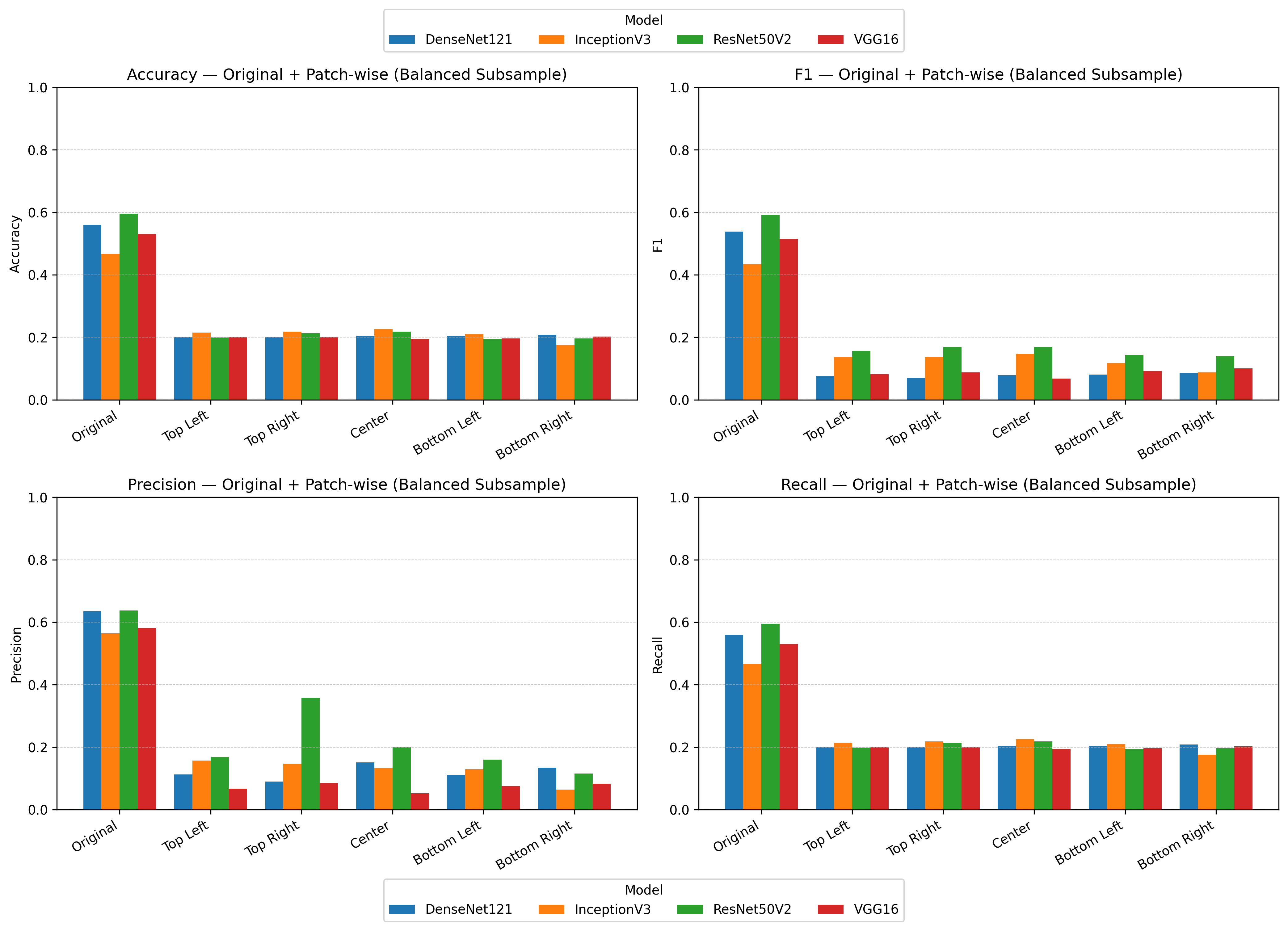}
\caption{\label{fig:cassava_bal} Performance of the four CNN architectures on the Cassava dataset. Classification accuracy (top left), macro-averaged F1 score (top right), macro-averaged precision (bottom left), and macro-averaged recall (bottom right) are reported for the original test images and the five background-dominated patch locations (top-left, top-right, center, bottom-left, and bottom-right).}
\end{figure}

The analyses presented thus far were conducted using the original test-set class distributions. Although the macro-averaged metrics reported in Section~\ref{subsec:class_sensitive} reduce the influence of majority classes during evaluation, they do not remove the imbalance present in the test data itself. To determine whether the contextual signal observed for the four high-bias datasets could be explained primarily by class imbalance, we repeated the evaluation using class-balanced test subsets constructed by randomly undersampling each class to the size of the smallest class. Figures~\ref{fig:weedcrop_bal}--\ref{fig:cassava_bal} summarize the resulting performance across the four CNN architectures.

Balancing the class distribution reduced contextual performance to different degrees across the evaluated datasets. DeepWeeds exhibited the largest change. Whereas the original evaluation showed mean patch accuracies substantially above the dataset-specific random-chance baseline, balancing the test set reduced patch accuracy to values close to the balanced nine-class chance level, with macro precision, recall, and F1 likewise remaining near their corresponding random baselines across all patch locations. These results indicate that class imbalance contributed substantially to the contextual signal previously observed for DeepWeeds.

A different pattern emerged for WeedCrop and Cassava. Although balancing reduced performance for several architectures and patch locations, above-chance classification remained evident for multiple experimental conditions. For WeedCrop, InceptionV3 and ResNet50V2 returned to the balanced binary chance level almost exactly — accuracy near 50\%, precision near 25\%, and F1 near 33\% — across every patch location. DenseNet121 and VGG16, in contrast, retained clearly above-chance performance after balancing, with precision reaching approximately 77–78\% and F1 reaching approximately 45–54\% across most spatial locations. For Cassava, ResNet50V2 retained the strongest contextual signal after balancing, with precision reaching approximately 36\% at select patch locations. InceptionV3 also remained marginally above the balanced chance level in terms of accuracy, whereas DenseNet121 and VGG16 returned to values close to or below the balanced five-class chance baseline across most evaluation metrics. Taken together, these results indicate that class imbalance explains part of the contextual signal identified in the previous sections but does not fully account for the above-chance performance observed for all datasets and architectures.

OpenSprayer exhibited the greatest robustness to balancing. Patch accuracies remained consistently above the balanced binary chance level across all evaluated architectures, with corresponding macro precision, recall, and F1 values also remaining substantially above their random baselines. Compared with the other high-bias datasets, balancing produced only a modest reduction in contextual performance, suggesting that the contextual information learned by the CNNs remained predictive even after controlling for class frequency.

Overall, balancing the class distribution reduced the contextual signal observed in several datasets, demonstrating that class imbalance contributes to the aggregate patch performance reported in Sections~\ref{sec:rq1}--\ref{subsec:class_sensitive}. However, the effect did not disappear uniformly across the evaluated datasets. Above-chance performance persisted for OpenSprayer and, to a lesser extent, for WeedCrop and Cassava, indicating that class imbalance alone does not fully explain the contextual reliance observed in this study. The remaining predictive information therefore suggests that CNNs learn additional contextual representations beyond those arising from class-frequency effects, motivating a closer examination of the features encoded within the learned representations.

\section{Discussion}
\label{discussion}

The experiments presented in this study address the two research questions introduced in Section~\ref{intro}. With respect to \textbf{RQ1}, CNNs trained for agricultural image classification achieved above-chance classification accuracy on background-dominated patches for six of the eight evaluated datasets, and substantially above chance for four of them. This finding holds despite the patches containing little to no information about the primary object of interest, indicating that contextual or acquisition-specific information contributed to model predictions for the majority of the datasets examined.

With respect to \textbf{RQ2}, the strength of this behavior varied considerably across datasets, architectures, and spatial locations, and did not survive uniformly under more rigorous evaluation. Macro-averaged metrics revealed that high patch accuracy was not always accompanied by comparably high class-wide discrimination: DeepWeeds and, to a lesser extent, Cassava exhibited accuracy substantially higher than macro recall or F1, indicating that correct patch predictions were concentrated within a subset of the label space rather than distributed across classes. Class-balanced evaluation further showed that the contextual signal for DeepWeeds, and for two of the four architectures evaluated on WeedCrop, was substantially attenuated once class-frequency effects were removed, indicating that class imbalance contributed materially to the accuracy-based evidence observed in Section~\ref{sec:rq1} for these cases. At the same time, above-chance performance persisted for OpenSprayer across all four architectures, and for a subset of architectures on WeedCrop and Cassava, after balancing. 
Class imbalance is therefore not a sufficient explanation for the contextual reliance observed in this study, but it is a substantial contributor for some datasets and architectures. Taken together, these findings indicate that contextual reliance is a genuine characteristic of CNN-based agricultural image classification, but one whose magnitude and underlying contributors differ substantially across datasets and architectures.

The persistence of above-chance performance after class-balanced evaluation is, on its own, informative but incomplete. It establishes that some datasets and architectures retain genuine class-discriminative information within background-dominated regions even after majority-class advantages are removed, but it does not indicate what visual information supports this discrimination. Accuracy, precision, recall, and F1 are behavioral measures: they quantify whether a model's predictions are correct, not which visual features the model relied upon to produce them. A model may achieve above-chance patch classification by learning genuinely predictive contextual regularities, such as characteristic soil color, canopy density, or acquisition-specific lighting associated with particular classes, or through more incidental correlations specific to how a given dataset was collected, and the evaluation protocol used in this study cannot distinguish between these possibilities. This limitation is consistent with the aggregate, behavioral nature of the metrics available at this stage of analysis, and it points to feature-level, rather than performance-level, inquiry as a necessary direction for further characterizing contextual bias in these CNNs.

These findings also strengthen the growing body of evidence suggesting that contextual reliance is not confined to a particular imaging domain. Previous studies have reported above-chance classification from background-only or context-only image regions for curated object datasets and cancer pathology images~\citep{martinez2026detection,okonoda2026unmasking}. The present work extends these observations to agricultural image classification while introducing a comparative evaluation across datasets with substantially different numbers of classes, acquisition conditions, and imaging scenarios. By incorporating dataset-specific random-chance baselines together with macro-averaged and class-balanced analyses, the present study further demonstrates that contextual reliance should not be interpreted solely through aggregate classification accuracy. Instead, its apparent strength depends both on the underlying dataset characteristics and on the evaluation protocol used to assess it.

\subsection{Limitations}
\label{subsec:limitations}

Several aspects of the present evaluation warrant caution when interpreting the results. First, patch extraction used a fixed absolute patch size relative to each architecture's input resolution rather than a size calibrated to object scale within each dataset; while this ensures a consistent proportion of the frame is sampled for a given architecture, the resulting patches are not guaranteed to be equally uninformative across all eight datasets, and images in which the primary object occupies a large fraction of the frame may yield corner or center patches containing partial object content despite the extraction procedure. In addition, the study considered four widely used CNN architectures but did not evaluate transformer-based vision models or foundation models, whose reliance on contextual information may differ. Second, the contextual bias categorization in Section~\ref{subsec:bias_category} relies on operational thresholds selected to summarize the observed distribution of margins in this benchmark rather than a threshold derived independently of the data. Third, the class-balanced evaluation in Section~\ref{subsec:balanced} used a single undersampling draw per dataset and condition rather than repeated resampling, and for datasets with few samples in the smallest class, the resulting balanced subsets may be small enough to introduce non-trivial variance into the reported balanced metrics; this variance was not separately quantified in the present study. Finally, the evaluation protocol used throughout this study is behavioral: it establishes whether background-dominated regions are class-predictive, but does not identify the specific visual features underlying this predictiveness, a limitation discussed further below.

\subsection{Future Work}
\label{sec:future_work}

The results of this study establish that contextual information remains predictive for several datasets even after accounting for class imbalance, but they do not reveal what visual information within the background-dominated regions supports these predictions. Understanding this remaining contextual signal requires moving beyond behavioral evaluation toward analysis of the internal feature representations learned by CNNs. Examining neuron activations and the semantic concepts encoded within hidden layers offers a natural next step for determining whether different levels of contextual reliance correspond to different learned visual representations. Such analyses would complement the empirical framework presented here by explaining not only whether contextual bias exists, but also how it emerges within the learned feature hierarchy.

\section{Conclusion}
\label{sec:conclusion}

This study investigated whether CNNs trained for agricultural image classification rely on contextual information beyond the primary object of interest by comparing classification performance on original images with performance on background-dominated patches across eight agricultural benchmark datasets and four widely used CNN architectures. Background-dominated patches were classified above dataset-specific random chance for six of the eight datasets, and substantially above chance for four of them, indicating that contextual reliance is a recurring, though not universal, characteristic of the evaluated CNNs and datasets.

Additional analyses showed that this behavior does not admit a single explanation. Macro-averaged evaluation demonstrated that contextual information was not distributed uniformly across classes, while class-balanced evaluation showed that class imbalance explained a substantial portion of the observed contextual signal for some datasets and architectures but not others. Above-chance contextual classification nevertheless persisted in several architecture--dataset combinations, indicating that class imbalance alone does not fully account for the observed behavior.

These findings provide both an empirical and methodological contribution. Empirically, they constitute the systematic evaluation across a substantially larger and more heterogeneous set of datasets than previously examined. Methodologically, by extending an evaluation protocol previously applied to curated object recognition and cancer pathology imaging, this work provides further evidence that contextual bias is a recurring property of CNN-based image classification rather than a phenomenon confined to a particular application domain. We therefore advocate complementing conventional benchmark evaluation with explicit assessments of contextual reliance, as demonstrated in this study, to obtain a more complete understanding of CNN behavior, including the visual information underlying contextual reliance that behavioral metrics alone cannot identify.

\section*{Acknowledgments}
The research was funded in part by National Science Foundation grant 2148878. 

During the preparation of this work, the author(s) used ChatGPT-4 in order to improve the readability and language of the paper. After using this tool/service, the author(s) reviewed and edited the content and take(s) full responsibility for the content of the publication.

\section*{Data Availability}
The datasets analyzed in this study are publicly available.
PlantVillage is available at \url{https://www.kaggle.com/datasets/emmarex/plantdisease/data}, PlantDoc is available at \url{https://www.kaggle.com/datasets/nirmalsankalana/plantdoc-dataset }, DeepWeeds is available at \url{https://github.com/AlexOlsen/DeepWeeds}, Cassava Disease is available at \url{https://www.kaggle.com/c/cassava-leaf-disease-classification}, WeedCrop is available at \url{ https://www.kaggle.com/datasets/vinayakshanawad/weedcrop-image-dataset}, Oxford 102 Flower is available at \url{https://www.robots.ox.ac.uk/~vgg/data/flowers/102/}, 
Sugarcane Disease Dataset is available at \url{https://data.mendeley.com/datasets/9twjtv92vk/1}, and
Open Sprayer Images is available at \url{ https://www.kaggle.com/datasets/vinayakshanawad/weedcrop-image-dataset}

\bibliographystyle{plain}
\bibliography{sample}
\end{document}